\documentclass[11pt,journal,draftcls,a4paper,onecolumn]{IEEEtran}

\usepackage[utf8]{inputenc}
\usepackage[T1]{fontenc}
\usepackage{amsmath}
\usepackage{amssymb}
\usepackage{mathrsfs}
\usepackage{color}
\usepackage{cite}
\usepackage{algorithm,algorithmic}
\usepackage{graphicx}
\usepackage{multirow}

\usepackage[labelformat=simple]{subcaption}

\title{Detecting Changes Between Optical Images of Different Spatial and Spectral Resolutions: a Fusion-Based Approach}

\author{Vinicius Ferraris, Nicolas Dobigeon, \IEEEmembership{Senior Member, IEEE}, \\Qi Wei, \IEEEmembership{Member, IEEE}, and Marie Chabert
\thanks{Part of this work has been supported by Coordenação de Aperfeiçoamento de Ensino Superior (CAPES), Brazil, and EU FP7 through the ERANETMED JC-WATER Program, MapInvPlnt Project ANR-15-NMED-0002-02.}
\thanks{V. Ferraris, N. Dobigeon and M. Chabert are with University of Toulouse, IRIT/INP-ENSEEIHT, France  (email: \{vinicius.ferraris, nicolas.dobigeon, marie.chabert\}@enseeiht.fr).}
\thanks{Q. Wei is with Department of Engineering, University of Cambridge, CB2 1PZ, Cambridge, UK (email: qw245@cam.ac.uk).}}

\begin{document}
\maketitle

\begin{abstract}

Change detection is one of the most challenging issues when analyzing remotely sensed images. Comparing several multi-date images acquired through the same kind of sensor is the most common scenario. Conversely, designing robust, flexible and scalable algorithms for change detection becomes even more challenging when the images have been acquired by two different kinds of sensors. This situation arises in case of emergency under critical constraints. This paper presents, to the best of authors' knowledge, the first strategy to deal with optical images characterized by dissimilar spatial and spectral resolutions. Typical considered scenarios include change detection between panchromatic or multispectral and hyperspectral images. The proposed strategy consists of a 3-step procedure: i) inferring a high spatial and spectral resolution image by fusion of the two observed images characterized one by a low spatial resolution and the other by a low spectral resolution, ii) predicting two images with respectively the same spatial and spectral resolutions as the observed images by degradation of the fused one and iii) implementing a decision rule to each pair of observed and predicted images characterized by the same spatial and spectral resolutions to identify changes. The performance of the proposed framework is evaluated on real images with simulated realistic changes.

\end{abstract}

\renewcommand{\thesection}{\Roman{section}}

\section{Introduction}

Change detection (CD) is one of the most investigated issues in remote sensing \cite{singhreview1989,riddcomparison1998,radkeimage2005,bovolotime2015}.
 %\cite{singhreview1989,nielsenmultivariate1998,riddcomparison1998,johnsonchange1998,bruzzonebayesian1999,bruzzonemrf1999,bruzzoneadaptive2002,ingladasimilarity2002,ingladachange2003,daddabbothree2004,luchange2004,radkeimage2005,albergaperformance2007,mercierconditional2007,nielsenregularized2007,nielsenkernel2011,liuunsupervised2012,dufusion2012,prendesnew2015,prendesperformance2015,bovolotime2015}.
As the name suggests, it consists in analyzing two or more multi-date (i.e., acquired at different time instants) images of the same scene to detect potential changes. Applications are diverse, from natural disaster monitoring to long-term tracking of urban and forest growth. Optical images have been the most studied remote sensing data for CD. They are generally well suited to map land-cover types at large scales  \cite{dalla_murachallenges2015}. Multi-band optical sensors use a spectral window with a particular width, often called spectral resolution, to sample part of the electromagnetic spectrum of the incoming light. The term \textit{spectral resolution} can also refer to the number of spectral bands and  multi-band images can be classified according to this number \cite{landgrebehyperspectral2002,campbellintroduction2011}. \textit{Panchromatic} (PAN) images are characterized by a low spectral resolution, sensing part of the electromagnetic spectrum with a single and generally wide spectral window. Conversely, \textit{multispectral} (MS) and \textit{hyperspectral} (HS) images have smaller spectral windows, allowing part of the spectrum to be sensed with higher precision. Multi-band optical imaging has become a very common modality of remote sensing, boosted by the advent of new finer spectral sensors \cite{colletmultivariate2006}. One of the major advantages of multi-band images is the possibility of detecting changes by exploiting not only the spatial but also the spectral information. There is no specific convention regarding the numbers of bands that characterize MS and HS images. Yet, MS images generally consists of a dozen of spectral bands while HS may have a lot more than a hundred. In complement to spectral resolution taxonomy, one may describe multi-band images in terms of their spatial resolution measured by the ground sampling interval (GSI), e.g. the distance, on the ground, between the center of two adjacent pixels \cite{dalla_murachallenges2015,elachiintroduction2006,campbellintroduction2011}. Informally, it represents the smallest object that can be resolved up to a specific pixel size. Then, the higher the resolution, the smaller the recognizable details on the ground: a \textit{high resolution} (HR) image has smaller GSI and finer details than a \textit{low resolution} (LR) one, where only coarse features are observable.
%Note that the pixel size in the image representation, by itself, \qw{does} not necessarily condition the spatial resolution of an image. It is possible to represent images with the same pixel size but with different spatial resolution. Indeed, it only denotes spatial resolution when it goes along with information about the instantaneous field of view (IFOV) pixel number.
Each image sensor is designed based on a particular signal-to-noise ratio (SNR). The reflected incoming light must be of sufficient energy to guarantee a sufficient SNR and thus a proper acquisition. To increase the energy level of the arriving signal, either the instantaneous field of view (IFOV) or the spectral window width must be increased. However these solutions are mutually exclusive. In other words, optical sensors suffer from an intrinsic energy trade-off that limits the possibility of acquiring images of both high spatial and high spectral resolutions \cite{pricespectral1997,elachiintroduction2006}. This trade-off prevents any simultaneous decrease of both the GSI and the spectral window width. Consequently, as an archetypal example, HS images are generally of lower spatial resolution than MS and PAN images.

Because of the common assumption of an additive Gaussian noise model for passive optical images, the most common CD techniques designed for single-band optical images are based on image differencing \cite{singhreview1989,riddcomparison1998,radkeimage2005,bovolotime2015}. When dealing with multi-band images, classical CD differencing methods have been adapted for such data through spectral change vectors \cite{riddcomparison1998,bovolotheoretical2007,bovoloframework2012} or transform analysis \cite{nielsenmultivariate1998,nielsenregularized2007}. Besides, most CD techniques assume that the multi-date images have been acquired by sensors of the same type \cite{bovolotime2015} with similar acquisition characteristics in terms of, e.g., angle-of-view, resolutions or noise model \cite{cantyautomatic2004,ingladapossibility2004}. Nevertheless, in some specific scenarios, for instance consecutive to natural disasters, such a constraint may not be ensured, e.g., images compatible with previously acquired ones may not be available in an acceptable timeframe. Such disadvantageous emergency situations yet require fast, flexible and accurate methods able to handle images acquired by sensors of different kinds \cite{ingladasimilarity2002,albergaperformance2007,mercierconditional2007,prendesnew2015,prendesperformance2015,prendeschange2015}. Facing with heterogeneity of data is a challenging task and must be carefully handled. However, since CD techniques for optical images generally rely on the assumption of data acquired by similar sensors, suboptimal strategies have been considered to make these techniques applicable when considering optical images of different spatial and spectral resolutions \cite{nielsenmultivariate1998,albergaperformance2007}. In particular, interpolation and resampling are classically used to obtain a pair of images with the same spatial and spectral resolutions \cite{albergaperformance2007,albergacomparison2007}. However, such a compromise solution may remain suboptimal since it considers each image individually without fully exploiting their joint characteristics and their complementarity. In this paper, we address the problem of unsupervised CD technique of multi-band optical images with different spatial and spectral resolutions. To the best of authors' knowledge, this is the first operational framework specifically designed to address this issue.
%Another important remark is that the majority of methods described for optical images are based strictly on the observed images without exploring the analysis of changes in the unknown latent scenes.

More precisely, this paper addresses the problem of CD between a pair of optical images acquired over the same scene at different time instants, one with low spatial and high spectral resolutions and one with high spatial and low spectral resolutions. The proposed approach consists in first fusing the two observed images. The result would be a high spatial and high spectral resolution image of the observed scene as if the two observed images were acquired at the same time or, in our case of study, if no change occurred between the two acquisition times. Otherwise, the result does not correspond to a truly observed scene but it contains the change information. The proposed fusion process explicitly relies on a physically-based sensing model which exploits the characteristics of the two sensors, following the frameworks in \cite{waldfusion1997,loncanhyperspectral2015}. These characteristics are subsequently resorted to obtain, by degradation of the fusion result, two so-called predicted images with the same resolutions as the observed images, i.e., one with low spatial resolution and high spectral resolutions and one with high spatial resolution and low spectral resolutions. In absence of any change, these two pairs of predicted and observed images should coincide, apart from residual fusion errors/inacurracies. Conversely, any change between the two observed images is expected to produce spatial and/or spectral alterations in the fusion result, which will be passed on the predicted images. Finally, each predicted image can be compared to the corresponding observed image of same resolution to identify possible changes. Since for each pair, the images to be compared are of the same resolution, classical CD methods dedicated to multi-band image can be considered \cite{radkeimage2005,bovolotime2015}. The final result is composed of two change detection maps with two different spatial resolutions.

%We propose an image restoration framework very related to DF \cite{weifast2015-2}. This will produce a single image with spectral resolution equal to the former and spatial resolution equal to the latter. This image represents the pseudolatent scene estimated from two different image modalities with the minimum overall reconstruction error for the considered pair. Some properties of the DF process and the ideas of a method introduced by \cite{waldfusion1997,loncanhyperspectral2015} allow to produce a pair of images from the restoration process with the same modality/characteristics of each observed image respectively. At this point two classical CD methods for homogeneous multi-band image \cite{radkeimage2005,bovolotime2015} can be applied for each pair of same modality multi-band images formed by an observed image and a result of restoration process at the pseudolatent image. This will generate two different

The paper is organized as follows. Section \ref{sec:CD} introduces the proposed change detection framework, which is composed of three main steps: fusion, prediction and decision. The first two steps are described in Section \ref{sec:fusion_prediction} which introduces the forward model underlying the observation process. Section \ref{sec:homCD}, dedicated to the third step, discusses three CD techniques operating on mono- and/or multi-band images of identical spatial and spectral resolutions. Experimental results are provided in Section \ref{sec:sim}, where a specific simulation protocol is detailed. These results demonstrate the efficiency of the proposed CD framework. Section \ref{sec:conclusion} concludes this paper.

\section{Proposed change detection framework}
\label{sec:CD}

Lets us denote $t_1$ and $t_2$ the times of acquisition for two multi-band optical images over the same scene of interest. Assume that the image acquired at time $t_1$ is a high spatial resolution PAN or MS (HR-PAN/MS) image denoted as $\mathbf{Y}_{\mathrm{HR}}^{t_1}\in \mathbb{R}^{n_{\lambda}\times n} $ and the one acquired at time $t_2$ is a low spatial resolution HS (LR-HS) image denoted as $\mathbf{Y}_{\mathrm{LR}}^{t_2} \in \mathbb{R}^{m_{\lambda}\times m}$, where
	\begin{itemize}
        \item $n = n_{\mathrm{r}}\times n_{\mathrm{c}}$ is the number of pixels in each band of the HR-PAN/MS image,
		\item $m = m_{\mathrm{r}}\times m_{\mathrm{c}}$ is the number of pixels in each band of the LR-HS image, with $m<n$,
		\item $n_{\lambda}$ is the number of bands in the HR-PAN/MS image,
		\item $m_{\lambda}$ is the number of bands in the LR-HS image, with $n_{\lambda}<m_{\lambda}$.
	\end{itemize}
The main difficulty which prevents any naive implementation of classical CD methods results from the differences in spatial and spectral resolutions of the two observed images, i.e., $m\neq n$ and $n_{\lambda}\neq m_{\lambda}$.

Besides, in digital image processing, it is common to consider the image formation process as a sequence of transformations of the original scene into an output image. The output image of a given sensor is thus a particular limited representation of the original scene with characteristics imposed by the processing pipeline of that sensor, called \textit{image signal processor} (ISP). The original scene cannot be exactly represented because of its continuous nature. Nevertheless, to represent the ISP pipeline as a sequence of transformations, it is usual to consider a very fine digital approximation of the scene representation as the input image. Following this paradigm, the two observed images $\mathbf{Y}_{\mathrm{HR}}^{t_1}$ and $\mathbf{Y}_{\mathrm{LR}}^{t_2}$ are assumed to be spectrally and spatially degraded versions of two corresponding latent (i.e., unobserved) high resolution hyperspectral images (HR-HS) $\mathbf{X}^{t_1}$ and $\mathbf{X}^{t_2}$, respectively,
\begin{equation}
\label{eq:forward_model_0}
		\begin{array}{rcl}
			\mathbf{Y}_{\mathrm{HR}}^{t_1} &=& T_{\mathrm{HR}}\left[\mathbf{X}^{t_1}\right]  \\
			\mathbf{Y}_{\mathrm{LR}}^{t_2} &=& T_{\mathrm{LR}}\left[\mathbf{X}^{t_2}\right]
		\end{array}
\end{equation}
where $T_{\mathrm{HR}}\left[\cdot\right]$ and $T_{\mathrm{LR}}\left[\cdot\right]$ stand for spectrally and spatially degradation operators and $\mathbf{X}^{t_j} \in \mathbb{R}^{m_{\lambda}\times n}$ ($j=1, 2$). Note that these two unobserved images $\mathbf{X}^{t_j} \in \mathbb{R}^{m_{\lambda}\times n}$ ($j=1, 2$) share the same spatial and spectral characteristics and, if they were available, they could be resorted as inputs of classical CD techniques operating on images of same resolutions.

When the two images $\mathbf{Y}_{\mathrm{HR}}^{t_1}$ and $\mathbf{Y}_{\mathrm{LR}}^{t_2}$ have been acquired at the same time, i.e., $t_1=t_2$, no change is expected and the latent images $\mathbf{X}^{t_1}$ and $\mathbf{X}^{t_2}$ should represent exactly the same scene, i.e., $\mathbf{X}^{t_1} = \mathbf{X}^{t_2} \triangleq \mathbf{X}$. In such a particular context, recovering an estimate $\hat{\mathbf{X}}$ of the HR-HS latent image $\mathbf{X}$ from the two degraded images $\mathbf{Y}_{\mathrm{HR}}^{t_1}$ and $\mathbf{Y}_{\mathrm{LR}}^{t_2}$ can be cast as a fusion problem, for which efficient methods have been recently proposed \cite{weihyperspectral2015,loncanhyperspectral2015,weifast2015-2,weibayesian2015-2}. Thus, in the case of a perfect fusion process, the no-change hypothesis $\mathcal{H}_0$ can be formulated as
\begin{equation}
\label{eq:forward_model_no_change}
\mathcal{H}_0 : \left\{
		\begin{array}{rcl}
			\mathbf{Y}_{\mathrm{HR}}^{t_1} &=& \hat{\mathbf{Y}}_{\mathrm{HR}}^{t_1}  \\
			\mathbf{Y}_{\mathrm{LR}}^{t_2} &=& \hat{\mathbf{Y}}_{\mathrm{LR}}^{t_2}
		\end{array}
        \right.
\end{equation}
where
\begin{equation}
\label{eq:predicted image}
		\begin{array}{rcl}
			 \hat{\mathbf{Y}}_{\mathrm{HR}}^{t_1} &\triangleq & T_{\mathrm{HR}}\left[\hat{\mathbf{X}}\right]  \\
			 \hat{\mathbf{Y}}_{\mathrm{LR}}^{t_2} &\triangleq & T_{\mathrm{LR}}\left[\hat{\mathbf{X}}\right]
		\end{array}
\end{equation}
are the two predicted HR-PAN/MS and LR-HS images from the estimated HR-HS latent image $\hat{\mathbf{X}}$.

\begin{figure}[h!]
	\centering
	\includegraphics[width=0.45\textwidth]{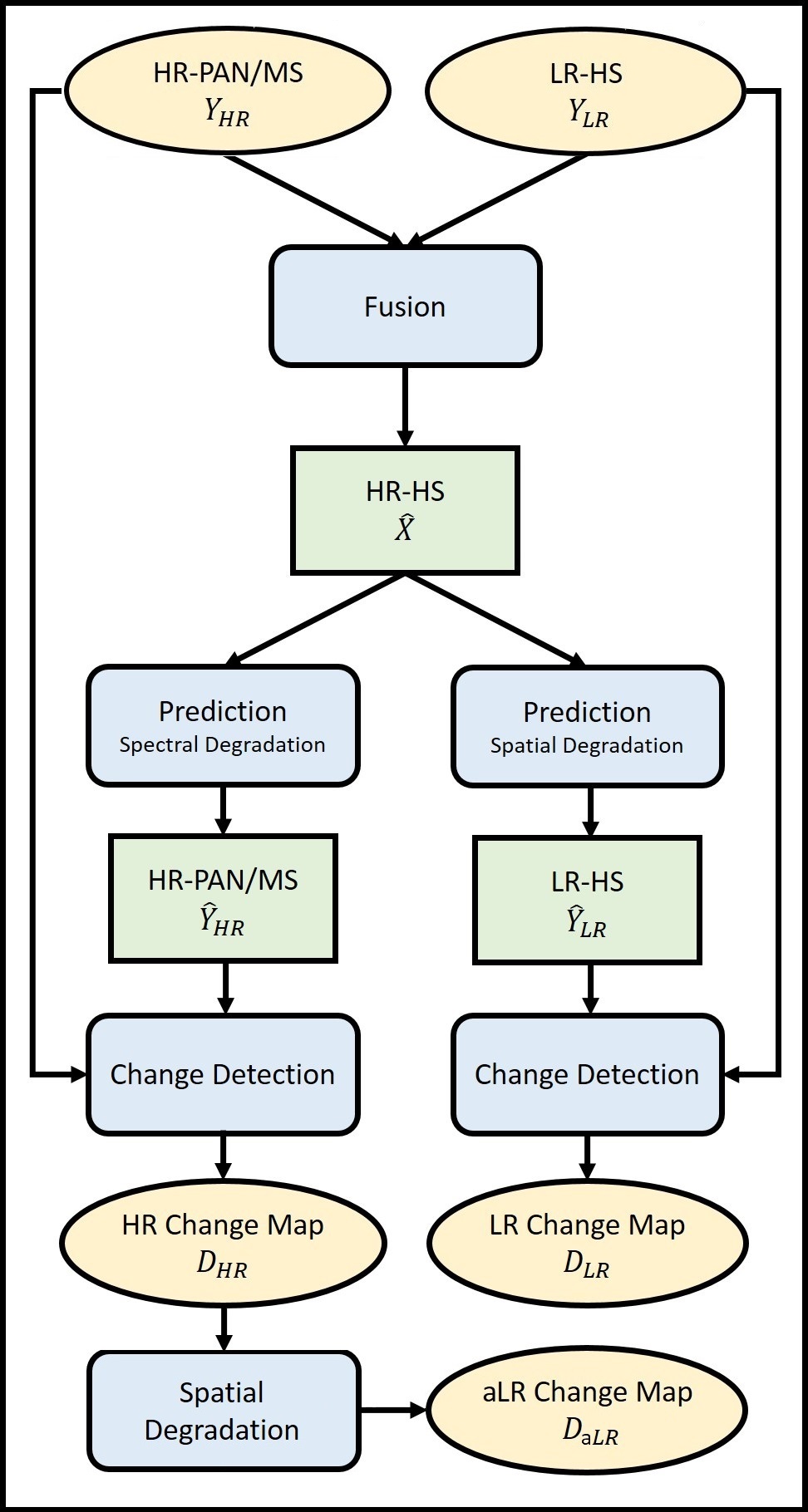}%
	\caption{Change detection framework.}%
	\label{fig:CDF}%
\end{figure}

When there exists a time interval between acquisitions, i.e. when $t_1\neq t_2$, a change may occur meanwhile. In this case, no common latent image $\mathbf{X}$ can be defined since $\mathbf{X}^{t_1} \neq \mathbf{X}^{t_2}$. However, since $\mathbf{X}^{t_1}$ and $\mathbf{X}^{t_2}$ represent the same area of interest, they are expected to keep a certain level of similarity. Thus, the fusion process does not lead to a common latent image, but to a pseudo-latent image $\hat{\mathbf{X}}$ from the observed image pair $\mathbf{Y}_{\mathrm{HR}}^{t_1}$ and $\mathbf{Y}_{\mathrm{LR}}^{t_2}$, which consists of the best joint approximation of latent images $\mathbf{X}^{t_1}$ and $\mathbf{X}^{t_2}$. Moreover, since $\hat{\mathbf{X}}\neq \mathbf{X}^{t_1}$ and $\hat{\mathbf{X}} \neq \mathbf{X}^{t_2}$, the forward model \eqref{eq:forward_model_0} does not hold to relate the pseudo-latent image $\hat{\mathbf{X}}$ to the observations $\mathbf{Y}_{\mathrm{HR}}^{t_1}$ and $\mathbf{Y}_{\mathrm{LR}}^{t_2}$. More precisely, when changes have occurred between the two time instants $t_1$ and $t_2$, the change hypothesis $\mathcal{H}_1$ can be stated as
\begin{equation}\label{eq:forward_model_change}
\mathcal{H}_1 : \left\{
		\begin{array}{rcl}
			\mathbf{Y}_{\mathrm{HR}}^{t_1} &\neq& \hat{\mathbf{Y}}_{\mathrm{HR}}^{t_1} \\
			\mathbf{Y}_{\mathrm{LR}}^{t_2} &\neq& \hat{\mathbf{Y}}_{\mathrm{LR}}^{t_2}.
		\end{array}
        \right.
\end{equation}
More precisely, both inequalities in \eqref{eq:forward_model_change} should be understood in a pixel-wise sense since any change occurring between $t_1$ and $t_2$ is expected to affect some spatial locations in the images. As a consequence, both diagnosis in \eqref{eq:forward_model_no_change} and \eqref{eq:forward_model_change} naturally induce pixel-wise rules to decide between the no-change and change hypothesis $\mathcal{H}_0$ and $\mathcal{H}_1$. This work specifically proposes to derive a CD technique able to operate on the two observed images $\mathbf{Y}_{\mathrm{HR}}^{t_1}$ and $\mathbf{Y}_{\mathrm{LR}}^{t_2}$. It mainly consists of a the following 3-steps, sketched in Fig. \ref{fig:CDF}
\begin{enumerate}
  \item \emph{fusion}: estimating the HR-HS pseudo-latent image $\hat{\mathbf{X}}$ from $\mathbf{Y}_{\mathrm{HR}}^{t_1}$ and $\mathbf{Y}_{\mathrm{LR}}^{t_2}$,
  \item \emph{prediction}: reconstructing the two HR-PAN/MS and LR-HS images $\hat{\mathbf{Y}}_{\mathrm{HR}}^{t_1}$ and $\hat{\mathbf{Y}}_{\mathrm{LR}}^{t_2}$ from  $\hat{\mathbf{X}}$,
  \item \emph{decision}: deriving HR and LR change maps $\hat{\mathbf{D}}_{\mathrm{HR}}$ and $\hat{\mathbf{D}}_{\mathrm{LR}}$ associated with the respective pairs of observed and predicted HR-PAN/MS and LR-HS images, namely,
    \begin{equation}
      \Upsilon_{\mathrm{HR}} = \left\{\mathbf{Y}_{\mathrm{HR}}^{t_1},\mathbf{\hat{Y}}_{\mathrm{HR}}^{t_1}\right\} \quad \text{and} \quad  \Upsilon_{\mathrm{LR}} =
      \left\{\mathbf{Y}_{\mathrm{LR}}^{t_2},\mathbf{\hat{Y}}_{\mathrm{LR}}^{t_2}\right\}.
    \end{equation}
    An alternate LR (aLR) change map, denoted as $\hat{\mathbf{D}}_{\mathrm{aLR}}$, is also computed by spatially degrading the HR change map $\hat{\mathbf{D}}_{\mathrm{HR}}$ with respect to the spatial operator $T_{\mathrm{LR}}\left[\cdot\right]$ and then comparing if at least one of the $\hat{\mathbf{D}}_{\mathrm{HR}}$ pixels associated to a given $\hat{\mathbf{D}}_{\mathrm{LR}}$ pixel leads to the same change/no-change decision.
\end{enumerate}

One should highlight the fact that this later \emph{decision} step only requires to implement CD techniques within two pairs of optical images $\Upsilon_{\mathrm{HR}}$ and $\Upsilon_{\mathrm{LR}}$ of same spatial and spectral resolutions, thus overcoming the initial issue raised by analyzing observed images $\mathbf{Y}_{\mathrm{HR}}^{t_1}$ and $\mathbf{Y}_{\mathrm{LR}}^{t_2}$ with dissimilar resolutions.

To establish the rationale underlying this framework, one may refer to the two main properties required by any fusion procedure: consistency and synthesis \cite{loncanhyperspectral2015}. The former one requires the reversibility of the fusion process: the original LR-HS and HR-PAN/MS can be obtained by proper degradations of the fused HR-HS image. The latter requires that the fused HR-HS image must be as similar as possible to the image of the same scene that would be obtained by sensor at the same resolution. Similarly, the generic framework proposed by Wald \emph{et al.} for fusion image quality assessment \cite{waldfusion1997} can also be properly stated by assigning the consistency and synthesis properties a greater scope.

Moreover, it is also worth noting that the proposed framework has been explicitly motivated by the specific scenario of detecting changes between LR-HS and HR-PAN/MS optical images. However, it may be applicable for any other CD scenario, provided that the two following assumptions hold: i) firstly, a latent image can be estimated from the two observed images and ii) secondly, the latent and predicted images can be related through known transformations.

Note finally that the modality-time order is not fixed, and without loss of generality, one may state either $t_1\leq t_2$ either $t_2\leq t_1$. Thus, to lighten the notations, without any ambiguity, the superscripts $t_1$ and $t_2$ will be omitted in the sequel of this paper. The three main steps of the proposed framework are described in the following sections.

\section{Fusion and prediction steps}
\label{sec:fusion_prediction}
This section describes the fusion and prediction steps involved in the proposed CD framework. Both intimately rely on the forward model introduced in what follows.

\subsection{Forward model}
\label{subsec:IR}
When dealing with optical images, the sequences of transformations $T_{\mathrm{HR}}\left[\cdot\right]$ and $T_{\mathrm{LR}}\left[\cdot\right]$ intrinsic to the sensors over the pseudo-latent images $\mathbf{X}$ in \eqref{eq:forward_model_0} are generally classified as spectral and spatial degradations. Spatial degradations are related to the spatial characteristics of the sensor, such as the sampling scheme and the optical transfer function. Spectral degradations, in the other hand, are relative to the sensitivity to wavelength and the spectral sampling. In this work, following widely admitted assumptions \cite{waldfusion1997,loncanhyperspectral2015}, these transformations are considered as linear degradations of the pseudo-latent image. Thus, benefiting from convenient matrix representations, the observed images can be expressed as
\begin{equation}
\label{eq:forward_model_1}
		\begin{array}{rcl}
			\mathbf{Y}_{\mathrm{HR}} &\approx& \mathbf{L}\mathbf{X},  \\
			\mathbf{Y}_{\mathrm{LR}} &\approx& \mathbf{X}\mathbf{R}.
		\end{array}
\end{equation}
The degradation resulting from the left multiplication by $\mathbf{L} \in \mathbb{R}^{n_{\lambda}\times m_{\lambda}}$ models the combination of some spectral bands for each pixel. This degradation corresponds to a spectral resolution reduction with respect to the pseudo-latent image $\mathbf{X}$ as in \cite{yokoyacross-calibration2013,weifast2015-2}. In practice, this degradation models an intrinsic characteristic of the sensor called spectral response. It can be either learned by cross-calibration or known \textit{a priori}.

Conversely, the right multiplication by $\mathbf{R} \in \mathbb{R}^{n\times m}$ degrades the pseudo-latent image by linear combinations of pixels within a given spectral band, thus reducing the spatial resolution. The right degradation matrix $\mathbf{R}$ may model the combination of various transformations which are specific of sensor architectures and take into account external factors such as wrap, blurring, translation, decimation, etc \cite{yokoyacross-calibration2013,heideflexisp:2014,weifast2015-2}. In this work, only space invariant blurring and decimation will be considered. Geometrical transformations such as wrap and translations can be corrected using image co-registration techniques in pre-processing steps. A space-invariant blur can be modeled by a symmetric convolution kernel, yielding a sparse symmetric Toeplitz matrix $\mathbf{B} \in \mathbb{R}^{n\times n}$ \cite{weihyperspectral2015}. It operates a cyclic convolution on the image bands individually. The decimation operation $\mathbf{S} \in \mathbb{R}^{n\times m}$ corresponds to a $d = d_{r} \times d_{c}$ uniform downsampling\footnote{The inverse downsampling transformation $\mathbf{S}^{T}$ represent an upsampling transformation by zero interpolation from $m$ to $n$.} operator with $ m = n/d$ ones on the block diagonal and zeros elsewhere, such that $\mathbf{S}^{T}\mathbf{S} = \mathbf{I}_{m}$ \cite{weifast2015-2}. Hence, the spatial degradation operation corresponds to the composition $\mathbf{R} = \mathbf{B}\mathbf{S} \in \mathbb{R}^{n\times m}$.

The approximating symbol $\approx$ in \eqref{eq:forward_model_1} stands for any mismodeling effects or acquisition noise, which is generally considered as additive and Gaussian \cite{bovolotime2015,elachiintroduction2006,weihyperspectral2015,loncanhyperspectral2015,weifast2015-2,weibayesian2015-2}. The full degradation model can thus be written as
\begin{equation}
        \label{eq:fusion_problem}
		\begin{array}{rcl}
			\mathbf{Y}_{\mathrm{HR}} &=& \mathbf{L}\mathbf{X} + \mathbf{N}_{\mathrm{HR}},  \\
			\mathbf{Y}_{\mathrm{LR}} &=& \mathbf{X}\mathbf{BS} + \mathbf{N}_{\mathrm{LR}}.
		\end{array}
\end{equation}
The additive noise matrices are assumed to be distributed according to matrix normal distributions\footnote{The probability density function, $p(\mathbf{X}|\mathbf{M},\mathbf{\Sigma}_{r},\mathbf{\Sigma}_{r})$ of a matrix normal distribution $\mathcal{M}\mathcal{N}_{r,c}(\mathbf{M},\mathbf{\Sigma}_{r},\mathbf{\Sigma}_{c})$ is given by
$$ p\left(\mathbf{X}|\mathbf{M},\mathbf{\Sigma}_{r},\mathbf{\Sigma}_{r}\right) = \frac{ \exp \left(-\frac{1}{2}tr \left[
\mathbf{\Sigma}_{c}^{-1} \left(\mathbf{X}-\mathbf{M}\right)^{T} \mathbf{\Sigma}_{r}^{-1} \left(\mathbf{X}-\mathbf{M}\right) \right]\right)}{\left(2\pi\right)^{rc/2}\left|\mathbf{\Sigma}_{c}\right|^{r/2}\left|\mathbf{\Sigma}_{r}\right|^{c/2}}$$
where $\mathbf{M} \in \mathbb{R}^{r\times c}$ is the mean matrix, $\mathbf{\Sigma}_{r} \in \mathbb{R}^{r\times r}$  is the row covariance matrix and $\mathbf{\Sigma}_{c} \in \mathbb{R}^{c\times c}$ is the column covariance matrix.} \cite{guptamatrix1999}, as follows
	\begin{equation*}
		\begin{array}{cl}
			\mathbf{N}_{\mathrm{HR}} \sim \mathcal{M}\mathcal{N}_{m_{\lambda},m}(\mathbf{0}_{m_{\lambda}\times m},\mathbf{\Lambda}_{\mathrm{HR}},\mathbf{I}_{m}), \\
			\mathbf{N}_{\mathrm{LR}} \sim \mathcal{M}\mathcal{N}_{n_{\lambda},n}(\mathbf{0}_{n_{\lambda}\times n},\mathbf{\Lambda}_{\mathrm{LR}},\mathbf{I}_{n}).
		\end{array}
	\end{equation*}
Note that the row covariance matrices $\mathbf{\Lambda}_{\mathrm{HR}}$ and $\mathbf{\Lambda}_{\mathrm{LR}}$ carry the information of the spectral variance in-between bands. Since the noise is spectrally colored, these matrices are not necessarily diagonal. In the other hand, since the noise is assumed spatially independent, the column covariance matrices correspond to identity matrices, e.g., $\mathbf{I}_{m}$ and $\mathbf{I}_{n}$. In real applications, since the row covariance matrices are an intrinsic characteristic of the sensor, they are estimated by a prior calibration \cite{yokoyacross-calibration2013}. In this paper, to reduce the number of unknown parameters we assume that $\mathbf{\Lambda}_{\mathrm{HR}}$ and $\mathbf{\Lambda}_{\mathrm{LR}}$ are both diagonal. This hypothesis implies that the noise is independent from one band to another and is characterized by a specific variance in each band \cite{weifast2015-2}.

\subsection{Fusion process}
\label{subsec:fusion}
The forward observation model \eqref{eq:fusion_problem} has been exploited in many applications involving optical multi-band images, specially those related to image restoration such as fusion and superresolution \cite{yokoyacross-calibration2013,weifast2015-2}. Whether the objective is to fuse multi-band images from different spatial and spectral resolutions or to increase the resolution of a single one, it consists in compensating the energy trade-off of optical multi-band sensors to get a higher spatial and spectral resolution image compared to the observed image set. One popular approach to conduct fusion consists in solving an inverse problem, formulated through the observation model. In the specific context of HS pansharpening (i.e., fusing PAN and HS images), such an approach has proven to provide the most reliable fused product, with a reasonable computational complexity \cite{loncanhyperspectral2015}. For these reasons, this is the strategy followed in this work and it is briefly sketched in what follows.

Because of the additive nature and the statistical properties of the noise $\mathbf{N}_{\mathrm{HR}}$ and $\mathbf{N}_{\mathrm{LR}}$, both observed images $\mathbf{Y}_{\mathrm{HR}}$ and $\mathbf{Y}_{\mathrm{LR}}$ are assumed to be distributed according to matrix normal distributions	
	\begin{equation}
		\begin{array}{cl}
			\mathbf{Y}_{\mathrm{HR}}|\mathbf{X} \sim \mathcal{M}\mathcal{N}_{m_{\lambda},m}(\mathbf{LX},\mathbf{\Lambda}_{\mathrm{HR}},\mathbf{I}_{m}) \\
			\mathbf{Y}_{\mathrm{LR}}|\mathbf{X} \sim \mathcal{M}\mathcal{N}_{n_{\lambda},n}(\mathbf{XBS},\mathbf{\Lambda}_{\mathrm{LR}},\mathbf{I}_{n})
		\end{array}
	\end{equation}
Since the noise can be reasonably assumed sensor-dependent, the observed images can be assumed statistically independent. Consequently the joint likelihood function of the statistical independent observed data can be written
	\begin{equation}
		p(\mathbf{Y}_{\mathrm{HR}},\mathbf{Y}_{\mathrm{LR}}|\mathbf{X}) = p(\mathbf{Y}_{\mathrm{HR}}|\mathbf{X})p(\mathbf{Y}_{\mathrm{LR}}|\mathbf{X})
	\end{equation}
and the negative log-likelihood, defined up to an additive constant, is
	\begin{equation}\label{eq:fusion_MLE}
		\begin{split}
			-\log p(\boldsymbol{\Psi}|\mathbf{X}) &= \frac{1}{2}\left\|\mathbf{\Lambda}_{\mathrm{HR}}^{-\frac{1}{2}} \left(\mathbf{Y}_{\mathrm{HR}} - \mathbf{LX} \right) \right\|_{F}^{2} \\
    &+ \frac{1}{2}\left\|\mathbf{\Lambda}_{\mathrm{LR}}^{-\frac{1}{2}} \left(\mathbf{Y}_{\mathrm{LR}} - \mathbf{XBS} \right) \right\|_{F}^{2}
	\end{split}
	\end{equation}
where  $\boldsymbol{\Psi} = \left\{\mathbf{Y}_{\mathrm{HR}},\mathbf{Y}_{\mathrm{LR}}\right\}$ denotes the set of observed images and $\left\|\cdot\right\|_{F}^{2}$ stands for the Frobenius norm.

%	An inverse problem can be also formulated as an estimation problem \cite{idierbayesian2008}. Having the conditional distributions $p(\mathbf{Y}_{\mathrm{HR}}|\mathbf{X})$ and $p(\mathbf{Y}_{\mathrm{LR}}|\mathbf{X})$, to estimate the joint likelihood function, $p(\mathbf{Y}_{\mathrm{HR}},\mathbf{Y}_{\mathrm{LR}}|\mathbf{X})$, one may assume statistical independence. Indeed, in practical scenarios, the images have been acquired by different sensors. 	
%Let $\Psi$ define the observable multi-band image . The negative joint log-likelihood function, $\mathcal{L}(\mathbf{X}|\Psi)$, is
%		\begin{equation}
%	\label{eq:likelihood}
%		\mathcal{L}(\mathbf{X}|\Psi) = -\log p(\Psi|\mathbf{X}) = - \log p(\mathbf{Y}_{\mathrm{HR}}|\mathbf{X}) - \log p(\mathbf{Y}_{\mathrm{LR}}|\mathbf{X})
%	\end{equation}
%	
%	Replacing in (\ref{eq:likelihood}) the conditional distributions of each component of the set accordingly, one can obtain
%	\begin{equation}
%		\begin{split}
%		-\log p(\Psi|\mathbf{X}) &= \frac{1}{2} tr\left[ \left(\mathbf{Y}_{\mathrm{HR}} - \mathbf{LX} \right)^{T} \mathbf{\Lambda}_{\mathrm{HR}}^{-1} \left(\mathbf{Y}_{\mathrm{HR}} - \mathbf{LX} \right) \right]\\
%		&+ \frac{1}{2} tr\left[ \left(\mathbf{Y}_{\mathrm{LR}} - \mathbf{XBS} \right)^{T} \mathbf{\Lambda}_{\mathrm{LR}}^{-1} \left(\mathbf{Y}_{\mathrm{LR}} - \mathbf{XBS} \right) \right]  + C
%		\end{split}
%	\end{equation}
%	$C$ is a constan. Since it does not involve the data directly, it will be omitted in the following, leading to:

Computing the maximum likelihood estimator $\mathbf{\hat{X}}_{\mathrm{ML}}$ of $\mathbf{X}$ from the observed image set $\boldsymbol{\Psi}$ consists in minimizing \eqref{eq:fusion_MLE}. The aforementioned derivation intents to solve a linear inverse problem which can be ill-posed or ill-conditioned, according to the properties of the matrices $\mathbf{B}$, $\mathbf{S}$ and $\mathbf{L}$ defining the forward model \eqref{eq:fusion_problem}. To overcome this issue, additional prior information can be included, setting the estimation problem into the Bayesian formalism \cite{idierbayesian2008}. Following a maximum a posteriori (MAP) estimation, recovering the estimated pseudo-latent image $\hat{\mathbf{X}}$ from the linear model \eqref{eq:fusion_problem} consists in minimizing the negative log-posterior
	\begin{equation}
	\label{eq:lnposmatch}
		\begin{split}
			\hat{\mathbf{X}} \in  \operatornamewithlimits{argmin}_{\mathbf{X} \in \mathbb{R}^{m_{\lambda}\times n}}  &\left\{ \frac{1}{2}\left\|\mathbf{\Lambda}_{\mathrm{HR}}^{-\frac{1}{2}} \left(\mathbf{Y}_{\mathrm{HR}} - \mathbf{LX} \right) \right\|_{F}^{2} \right.\\
&\left.+ \frac{1}{2}\left\|\mathbf{\Lambda}_{\mathrm{LR}}^{-\frac{1}{2}} \left(\mathbf{Y}_{\mathrm{LR}} - \mathbf{XBS} \right) \right\|_{F}^{2} + \lambda\phi(\mathbf{X}) \right\}
		\end{split}
	\end{equation}
where $\phi(\cdot)$ defines an appropriate regularizer derived from the prior distribution assigned to $\mathbf{X}$ and $\lambda$ is a parameter that tunes the relative importance of the regularization and data terms. Computing the MAP estimator \eqref{eq:lnposmatch} is expected to provide the best approximation $\hat{\mathbf{X}}$ with the minimum distance to the latent images $\mathbf{X}^{t_1}$ and $\mathbf{X}^{t_2}$ simultaneously. This optimization problem is challenging because of the high dimensionality of the data $\mathbf{X}$. Nevertheless, Wei \emph{et al.}\cite{weifast2015-2} has proved that its solution can be efficiently computed for various relevant regularization terms $\phi(\mathbf{X})$. In this work, a Gaussian prior is considered, since it provides an interesting trade-off between accuracy and computational complexity, as reported in \cite{loncanhyperspectral2015}.

\subsection{Prediction}

The prediction step relies on the forward model \eqref{eq:fusion_problem} proposed in Section~\ref{subsec:IR}. As suggested by \eqref{eq:predicted image}, it merely consists in applying the respective spectral and spatial degradations to the estimated pseudo-latent image $\hat{\mathbf{X}}$, leading to
\begin{equation}
\label{eq:predicted image_2}
\begin{array}{rcl}
			 \hat{\mathbf{Y}}_{\mathrm{HR}} &= & \mathbf{L}\hat{\mathbf{X}}\\
			 \hat{\mathbf{Y}}_{\mathrm{LR}} &= & \hat{\mathbf{X}}\mathbf{BS}.
		\end{array}
\end{equation}

\section{Optical Image Homogeneous Change Detection}
\label{sec:homCD}

	This section presents the third and last step of the proposed CD framework, which consists in implementing decision rules to identify possible changes between the images composing the two pairs $\Upsilon_{\mathrm{HR}}=\left\{\mathbf{Y}_{\mathrm{HR}},\mathbf{\hat{Y}}_{\mathrm{HR}}\right\}$ and $\Upsilon_{\mathrm{LR}}=\left\{\mathbf{Y}_{\mathrm{LR}},\mathbf{\hat{Y}}_{\mathrm{LR}}\right\}$. As noticed in Section \ref{sec:CD}, these CD techniques operate on observed $\mathbf{Y}_{\cdot\mathrm{R}}$ and predicted $\hat{\mathbf{Y}}_{\cdot\mathrm{R}}$ images of same spatial and spectral resolutions, with $\cdot\in\left\{\mathrm{H},\mathrm{L}\right\}$, as in \cite{riddcomparison1998,johnsonchange1998,daddabbothree2004,radkeimage2005}. Unless explicitly specified, they can be employed whatever the number of bands. As a consequence, $\mathbf{Y}_{\cdot\mathrm{R}}$ and $\hat{\mathbf{Y}}_{\cdot\mathrm{R}}$ could refer to either PAN, MS or HS images and the two resulting CD maps are either of HR, either of LR, associated with the pairs $\Upsilon_{\mathrm{HR}}$ and $\Upsilon_{\mathrm{LR}}$, respectively. To lighten the notations, without loss of generality, in what follows, the pairs $\mathbf{Y}_{\cdot\mathrm{R}}$ and $\hat{\mathbf{Y}}_{\cdot\mathrm{R}}$ will be denoted $\mathbf{Y}_1\in \mathbb{R}^{\ell\times\eta}$ and $\mathbf{Y}_2 \in \mathbb{R}^{\ell\times\eta}$, which can be set as
%\begin{itemize}
%  \item $\mathbf{Y}_1={\mathbf{Y}}_{\mathrm{HR}}$ and $\mathbf{Y}_2=\hat{\mathbf{Y}}_{\mathrm{HR}}$ to obtain the HR CD binary map $\mathbf{D}_{\mathrm{HR}}$,
%  \item $\mathbf{Y}_1={\mathbf{Y}}_{\mathrm{LR}}$ and $\mathbf{Y}_2=\hat{\mathbf{Y}}_{\mathrm{LR}}$ to obtain the LR CD binary map $\mathbf{D}_{\mathrm{LR}}$,
%\end{itemize}
\begin{itemize}
  \item $\left\{\mathbf{Y}_1, \mathbf{Y}_2\right\}= \Upsilon_{\mathrm{LR}}$ to derive the estimated CD binary map $\hat{\mathbf{D}}_{\mathrm{LR}}$ at LR,
  \item $\left\{\mathbf{Y}_1, \mathbf{Y}_2\right\}= \Upsilon_{\mathrm{HR}}$ to derive the estimated CD binary map $\hat{\mathbf{D}}_{\mathrm{HR}}$ at HR and its spatially degraded aLR counterpart $\hat{\mathbf{D}}_{\mathrm{aLR}}$.
\end{itemize}
In this seek of generality, the numbers of bands and pixels are denoted $\ell$ and $\eta$, respectively. The spectral dimension $\ell$ depends on the considered image sets $\Upsilon_{\mathrm{HR}}$ or $\Upsilon_{\mathrm{LR}}$, i.e., $\ell=n_{\lambda}$ and $\ell=m_{\lambda}$ for HR and LR images, respectively\footnote{Note, in particular, that $\ell=n_{\lambda}=1$ when the set of HR images are  PAN images.}. Similarly, the spatial resolution of the CD binary map generically denoted as $\hat{\mathbf{D}}\in\mathbb{R}^\eta$ depends on the considered set of images $\Upsilon_{\mathrm{HR}}$ or $\Upsilon_{\mathrm{LR}}$, i.e., $\eta=n$ and $\eta=m$ for HR and LR images, respectively.

Three efficient CD techniques operating on images of same spatial and spectral resolutions are discussed below.

\subsection{Change vector analysis (CVA)}
\label{subsubsec:CVA}
When considering multi-band optical images after atmospheric and geometric pre-calibration, for a pixel at spatial location $p = (i_p,j_p)$, one may consider that
	\begin{equation}
		\begin{array}{cl}
			\mathbf{Y}_{1}(p) \sim \mathcal{N}(\boldsymbol{\mu}_1,\boldsymbol{\Sigma}_1)\\
			\mathbf{Y}_{2}(p) \sim \mathcal{N}(\boldsymbol{\mu}_2,\boldsymbol{\Sigma}_2)
		\end{array}
	\end{equation}
where $\boldsymbol{\mu}_1\in \mathbb{R}^{\ell}$ and $\boldsymbol{\mu}_2 \in \mathbb{R}^{\ell}$ correspond to the pixel spectral mean and $\boldsymbol{\Sigma}_1 \in \mathbb{R}^{\ell \times \ell}$ and $\boldsymbol{\Sigma}_2 \in \mathbb{R}^{\ell \times \ell}$ are the spectral covariance matrices (here they were obtained using the maximum likelihood estimator). The spectral change vector is defined by the squared Mahalanobis distance between the two pixels which can be computed from the pixel-wise spectral difference operator $\Delta\mathbf{Y}(p) = \mathbf{Y}_{1}(p) - \mathbf{Y}_{2}(p)$, i.e.,
	\begin{equation}
	\label{eq:MD}
		V_{\mathrm{CVA}}(p) = \left\|\Delta\mathbf{Y}(p)\right\|_{\mathbf{\Sigma}^{-1}}^{2} = \Delta\mathbf{Y}(p)^T\mathbf{\Sigma}^{-1}\Delta\mathbf{Y}(p)
	\end{equation}
where $\boldsymbol{\Sigma} = \boldsymbol{\Sigma}_1 + \boldsymbol{\Sigma}_2$. For a given threshold $\tau$, the pixel-wise statistical test can be formulated as
\begin{equation}
  V_{\mathrm{CVA}}(p) \overset{\mathcal{H}_1}{\underset{\mathcal{H}_0}{\gtrless}} \tau
\end{equation}
and the final CD map, denoted $\hat{\mathbf{D}}_{\mathrm{CVA}} \in \{0,1\}^\eta$ can be derived as
	\begin{equation}
	\label{eq:CVArule}
 \hat{D}_{\mathrm{CVA}}(p) = \left\{\begin{array}{lll}
             1 & \mbox{if } V_{\mathrm{CVA}}(p) \geq \tau & (\mathcal{H}_1)\\
			 0 & \mbox{otherwise}          & (\mathcal{H}_0).
				\end{array}\right.
\end{equation}
For a pixel which has not been affected by a change (hypothesis $\mathcal{H}_0$), the spectral difference operator is expected to be statistically described by $\Delta\mathbf{Y}(p) \sim \mathcal{N}(\boldsymbol{0},\mathbf{\Sigma})$. As a consequence, the threshold $\tau$ can be related to the probability of false alarm (PFA) of the test
\begin{eqnarray}
\label{eq:PFA}
P_\mathrm{FA} & = & \mathbb{P}\left[V_{\mathrm{CVA}}(p) > \tau \bigg| \mathcal{H}_0\right]
\end{eqnarray}
or equivalently,
\begin{eqnarray}
\label{eq:threshold}
\tau = \mathrm{F}_{\chi^2_{\ell}}^{-1}\left(1 - P_\mathrm{FA}\right)
\end{eqnarray}
where $\mathrm{F}_{\chi^2_{\ell}}^{-1}(\cdot)$ is the inverse cumulative
distribution function of the ${\chi^2_{\ell}}$ distribution.

\subsection{Spatially regularized change vector analysis}

Since CVA in its simplest form as presented in Section \ref{subsubsec:CVA} is a pixel-wise procedure, it significantly suffers from low robustness with respect to noise. To overcome this limitation, spatial information can be exploited by considering the neighborhood of a pixel to compute the final distance criterion, which is expected to make the change map spatially smoother. Indeed, changed pixels are generally gathered together into regions or clusters, which means that there is a high probability to observe changes in the neighborhood of an identified changed pixel \cite{radkeimage2005}. Let $\Omega_{p}^{L}$ denote the set of indexes of neighboring spatial locations of a given pixel $p$ defined by a surrounding regular window of size $L$ centered on $p$. The spatially smoothed energy map $\mathbf{V}_{\mathrm{sCVA}}$ of the spectral difference operator can be derived from its pixel-wise counterpart $\mathbf{V}_{\mathrm{CVA}}$ defined by \eqref{eq:MD} as
	\begin{equation}
		{V}_{\mathrm{sCVA}}(p) = \frac{1}{|\Omega_{p}^{L}|}\sum_{k \in \Omega_{p}^{L}}\omega(k){V}_{\mathrm{CVA}}(k)
	\end{equation}
where the weights $\omega(k) \in \mathbb{R}^{|\Omega_{p}^{L}|}$	implicitly define a spatial smoothing filter. In this work, they are chosen as $\omega(k)=1$, $\forall k \in\left\{1,\ldots, |\Omega_{p}^{L}|\right\}$.
Then, a decision rule similar to \eqref{eq:CVArule} can be followed to derive the final CD map $\hat{\mathbf{D}}_{\mathrm{sCVA}}$. Note, the choice of window size $L$ is based on the strong hypothesis of the window homogeneity. This choice thus may depend upon the kind of observed scenes.

\subsection{Iteratively-reweighted multivariate alteration detection (IR-MAD)}

The multivariate alteration detection (MAD) technique introduced in \cite{nielsenmultivariate1998} has been shown to be a robust CD method due to being well suited for analyzing multi-band image pair $\left\{\mathbf{Y}_1,\mathbf{Y}_2\right\}$ with possible different intensity levels. Similarly to the CVA and sCVA methods, it exploits an image differencing operator while better concentrating information related to changes into auxiliary variables. More precisely, the MAD variate is defined as $\Delta \tilde{\mathbf{Y}}(p) = \tilde{\mathbf{Y}}_1(p)-\tilde{\mathbf{Y}}_2(p)$ with
\begin{equation}
\begin{aligned}
  \tilde{\mathbf{Y}}_1(p) &= \mathbf{U}{\mathbf{Y}}_1(p)\\
  \tilde{\mathbf{Y}}_2(p) &= \mathbf{V}{\mathbf{Y}}_2(p)
\end{aligned}
\end{equation}
where $\mathbf{U}=\left[\mathbf{u}_{\ell}, \mathbf{u}_{\ell-1},\ldots,\mathbf{u}_{1}  \right]^T$ is a $\ell \times \ell$-matrix  composed of the $\ell\times 1$-vectors $\mathbf{u}_{j}$ identified by canonical correlation analysis and $\mathbf{V}=\left[\mathbf{v}_{\ell}, \mathbf{v}_{\ell-1},\ldots,\mathbf{v}_{1}  \right]^T$ is defined similarly. As in Equation (\ref{eq:MD}), the MAD-based change energy map can then be derived as
\begin{equation*}
  {V}_{\mathrm{MAD}}(p) = \|\Delta \tilde{\mathbf{Y}}(p)\|^2_{\boldsymbol{\Lambda}^{-1}}
\end{equation*}
where $\boldsymbol{\Lambda}$ is the diagonal covariance matrix of the MAD variates. Finally, the MAD CD map $\hat{\mathbf{D}}_{\mathrm{MAD}}$ can be pixel-wisely computed using a decision rule similar to \eqref{eq:CVArule} with a threshold $\tau$ related to the PFA by \eqref{eq:threshold}. In this work, the iteratively re-weighted version of MAD (IR-MAD) has been considered to better separate the change pixels from the no-change pixels \cite{nielsenregularized2007}.

\section{Experiments}
\label{sec:sim}

This section assesses the performance of the proposed fusion-based CD framework. First, the simulation protocol is described in Section \ref{subsec:SF}. Then, Section \ref{subsec:results} reports qualitative and quantitative results when detecting changes between HS and MS or PAN images.

\subsection{Simulation protocol}
\label{subsec:SF}

Evaluating performances of CD algorithms requires image pairs with particular characteristics, which makes them rarely freely available. Indeed, CD algorithms require images acquired at two different dates, presenting changes, geometrically and radiometrically pre-corrected and, for the specific problem addressed in this paper, coming from different optical sensor modes. Moreover, these image pairs need to be accompanied by ground-truth information in the form of validated CD mask.

To overcome this issue, this paper proposes to follow a strategy inspired by the protocol introduced in \cite{waldfusion1997} to assess the performance of pansharpening algorithms. This protocol relies on a unique reference HS image $\mathbf{X}_{\mathrm{ref}}$, also considered as HR. It avoids the need of  co-registered and geometrically corrected images by generating a pair of synthetic but realistic HR-PAN/MS and LR-HS images from this reference image and by including changes within a semantic description of this HR-HS image. In this work, this description is derived by spectral unmixing \cite{bioucas-diashyperspectral2013} and the full proposed protocol can be summarized as follows:
	\begin{enumerate}
		\item[i)] Given an HR-HS reference image $\mathbf{X}_{\mathrm{ref}} \in \mathbb{R}^{m_{\lambda} \times n}$, conduct linear unmixing to extract $K$ endmember signatures $\mathbf{M}^{t_1} \in \mathbb{R}^{m_{\lambda} \times K}$ and the associated abundance matrix $\mathbf{A}^{t_1} \in \mathbb{R}^{K \times n}$.
        \item[ii)] Define the HR-HS latent image $\mathbf{X}^{t_1}$ before change as
        \begin{equation}
  \mathbf{X}^{t_1} =  \mathbf{M}^{t_1}\mathbf{A}^{t_1}.
\end{equation}
		\item[iii)] Define a reference HR change mask $\mathbf{D}_{\mathrm{HR}}$ by selecting particular regions (i.e., pixels) in the HR-HS latent image $\mathbf{X}^{t_1}$ where changes occur. The corresponding LR change mask $\mathbf{D}_{\mathrm{LR}}$ is computed according to the spatial degradations relating the two modalities. Both change masks will be considered as the ground truth and will be compared to the estimated CD HR map $\hat{\mathbf{D}}_{\mathrm{HR}}$ and LR maps $\hat{\mathbf{D}}_{\mathrm{LR}}$ and $\hat{\mathbf{D}}_{\mathrm{aLR}}$, respectively, to evaluate the performance of the proposed CD technique.
        \item[iv)] According to this reference HR change mask, implement realistic change rules on the reference abundances $\mathbf{A}^{t_1}$ associated with pixels affected by changes. Several change rules applied to the reference abundance will be discussed in Section \ref{subsubsec:change_rules}. Note that theses rules may also require the use of additional endmembers that are not initially present in the latent image $\mathbf{X}^{t_1}$. The abundance and endmember matrices after changes are denoted as $\mathbf{A}^{t_2}$ and $\mathbf{M}^{t_2}$, respectively.
            %This change may imply the used of an additional $(K+1)$th endmembers, (\ref{eq:abd}). The abundances of a pixel has two important characteristic that must be maintained whatever chosen change rule \cite{bioucas-diashyperspectral2013}. If new endmembers are introduced, they must be added in $\mathbf{M}_{ref}$. The new abundances and endmembers matrices are called respectively $\mathbf{A}_{ch}$ and $\mathbf{M}_{ch}$.
		\item[v)] Define the HR-HS latent image $\mathbf{X}^{t_2}$ after changes by linear mixing such that
\begin{equation}
  \mathbf{X}^{t_2} =  \mathbf{M}^{t_2}\mathbf{A}^{t_2}.
\end{equation}
		\item[vi)] Generate a simulated observed HR-PAN/MS image $\mathbf{Y}_{\mathrm{HR}}$ by applying the spectral degradation $T_{\mathrm{HR}}\left[\cdot \right]$ either to the  before-change HR-HS latent image $\mathbf{X}^{t_1}$, either to the  after-change HR-HS latent image $\mathbf{X}^{t_2}$.
		\item[vii)] Conversely, generate a simulated observed LR-HS image $\mathbf{Y}_{\mathrm{LR}}$ by applying the spatial degradation $T_{\mathrm{LR}}\left[\cdot \right]$ either to the  after-change HR-HS latent image $\mathbf{X}^{t_2}$, or to the  before-change HR-HS latent image $\mathbf{X}^{t_1}$.
		\end{enumerate}	

\begin{figure}[h!]
	\centering
	\includegraphics[width=0.45\textwidth]{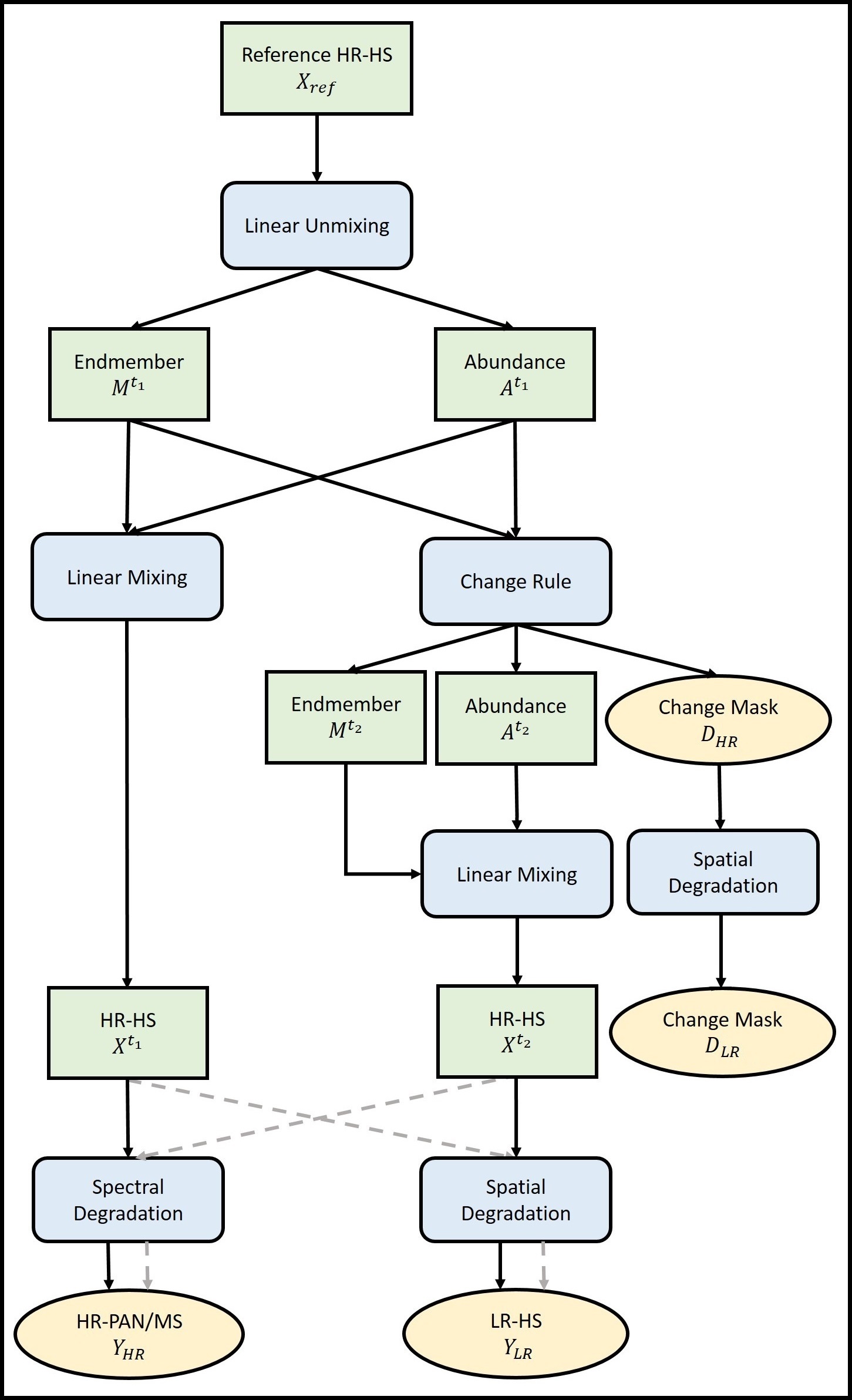}%
	\caption{Simulation protocol: two HR-HS latent images $\mathbf{X}^{t_1}$ (before changes) and $\mathbf{X}^{t_2}$ (after changes) are generated from the reference image. In temporal configuration 1 (black), the observed HR-PAN/MS image $\mathbf{Y}_{\mathrm{HR}}$ is a spectrally degraded version of $\mathbf{X}^{t_1}$ while the observed LR-HS image $\mathbf{Y}_{\mathrm{LR}}^{t_2}$ is a spatially degraded version of $\mathbf{X}^{t_2}$. In temporal configuration 2 (grey dashed lines),  the degraded images are generated from reciprocal HR-HS images.}
	\label{fig:CSF}%
\end{figure}

This protocol is illustrated in Fig. \ref{fig:CSF} and complementary information regarding these steps is provided in the following paragraphs.

\subsubsection{Reference image}

\label{subsubsec:reference_image}
The HR-HS reference image used in the simulation protocol is a $610 \times 330 \times 115$ HS image of the Pavia University in Italy acquired by the reflective optics system imaging spectrometer (ROSIS) sensor. A pre-correction has been conducted to smooth the atmospheric effects due to vapor water absorption by removing corresponding spectral bands. Then the final HR-HS reference image is of size $610 \times 330 \times 93$.

\subsubsection{Generating the HR-HS latent images: unmixing, change mask and change rules}
\label{subsubsec:change_rules}

To produce the HR-HS latent image $\mathbf{X}^{t_1}$ before change, the reference image $\mathbf{X}_{\mathrm{ref}}$ has been linearly unmixed, which provides the endmember matrix $\mathbf{M}^{t_1} \in \mathbb{R}^{m_{\lambda} \times K}$ and the matrix of abundances $\mathbf{A}^{t_1} \in \mathbb{R}^{K \times n}$ where $K$ is the number of endmembers. This number $K$ can be obtained by investigating the dimension of the signal subspace, for instance by conducting principal component analysis \cite{bioucas-diashyperspectral2013}. In this work, the linear unmixing has been conducted by coupling the vertex component analysis (VCA) \cite{nascimentovertex2005} as an endmember extraction algorithm and the fully constrained least squares (FCLS) algorithm \cite{Heinz2001} to obtain $\mathbf{M}^{t_1}$ and $\mathbf{A}^{t_1}$, respectively.

Given the HR-HS latent image $\mathbf{X}^{t_1}=\mathbf{M}^{t_1}\mathbf{A}^{t_1}$, the HR change mask $\mathbf{D}_{\mathrm{HR}}$ has been produced by selecting spatial regions in the HR-HS image affected by changes. This selection can be made randomly or by using prior knowledge on the scene. In this work, manual selection is performed.

\begin{figure}[h!]
\centering
			\begin{subfigure}[b]{0.24\textwidth}
					\centering	
					\includegraphics[trim={3.08cm 2.35cm 3.08cm 1cm},clip,scale=0.255]{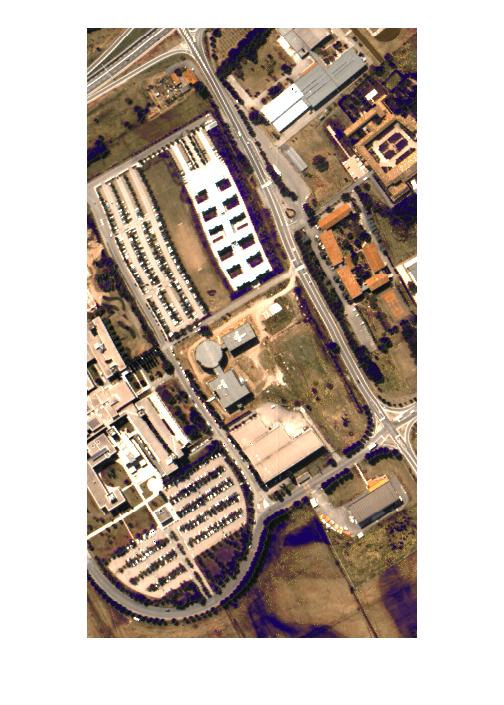}
					\caption{}
					\label{fig:cref}
			\end{subfigure}
			\begin{subfigure}[b]{0.24\textwidth}
					\centering	
					\includegraphics[trim={3.08cm 2.35cm 3.08cm 1cm},clip,scale=0.255]{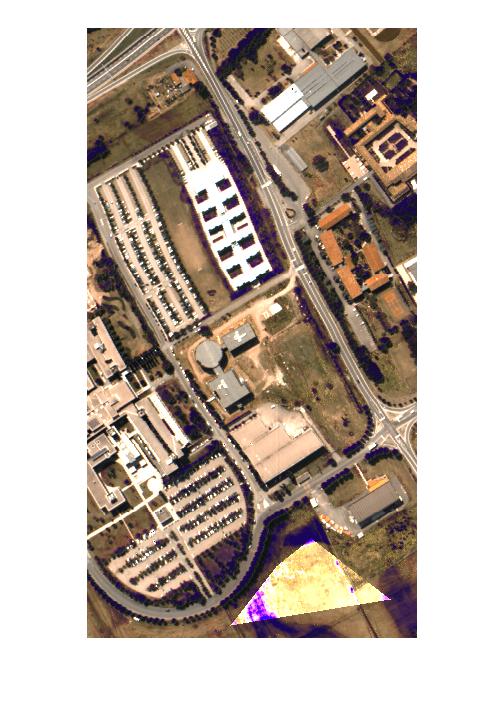}
					\caption{}
					\label{fig:chZero}
			\end{subfigure}
			\begin{subfigure}[b]{0.24\textwidth}
					\centering	
					\includegraphics[trim={3.08cm 2.35cm 3.08cm 1cm},clip,scale=0.255]{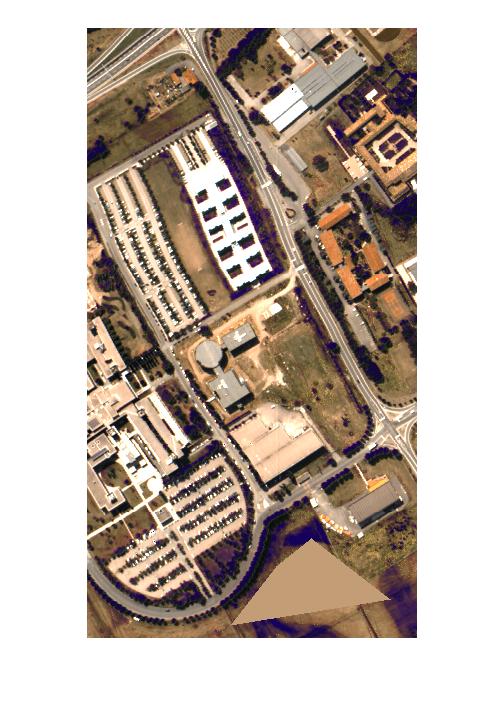}
					\caption{}
					\label{fig:chSame}
			\end{subfigure}
			\begin{subfigure}[b]{0.24\textwidth}
					\centering	
					\includegraphics[trim={3.08cm 2.35cm 3.08cm 1cm},clip,scale=0.255]{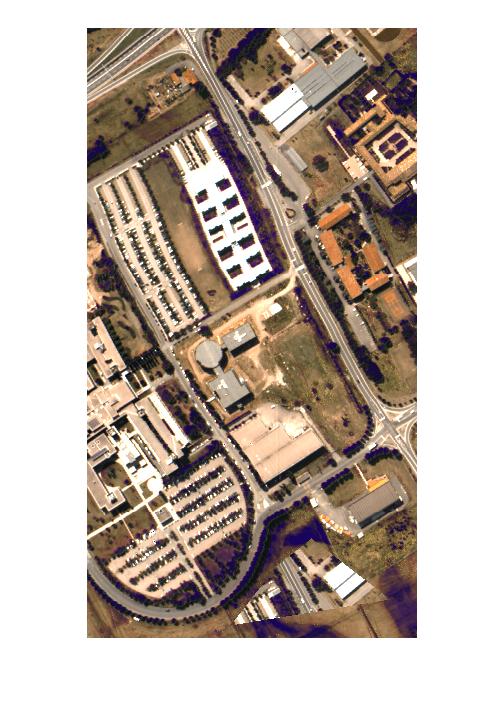}
					\caption{}
					\label{fig:chBlock}
			\end{subfigure}
			\caption{Change rules applied to the reference image \protect\subref{fig:cref}: \protect\subref{fig:chZero} zero-abundance, \protect\subref{fig:chSame} same abundance and \protect\subref{fig:chBlock} block abundance.}%
		\label{fig:CH}%
\end{figure}

Then, the change rules applied to the abundance matrix $\mathbf{A}^{t_1}$ to obtain the changed abundance matrix $\mathbf{A}^{t_2}$ are chosen such that they satisfy the standard positivity and sum-to-one constraints
		\begin{equation}
	\label{eq:abd}		
		\begin{aligned}
			&\text{Nonnegativity  } a_{k}^{t_2}(p) \geq 0 , \forall p\in \left\{1, \ldots, n\right\},  \forall k\in\left\{1, \ldots, K\right\}\\
			&\text{Sum-to-one    } \sum_{k=1}^{K} a_k^{t_2}(p) = 1,  \forall p\in \left\{1, \ldots, n\right\}
		\end{aligned}
	  \end{equation}
More precisely, three distinct change rules has been considered
	\begin{itemize}
		\item Zero abundance: find the most present endmember in the selected region, set all corresponding abundances to zero and rescale abundances associated with remaining endmembers in order to fulfill \eqref{eq:abd}. This change can be interpreted as a brutal disappearing of the most present endmember.
		\item Same abundance: choose a pixel abundance vector at random spatial location, set all abundance vectors inside the region affected by changes to the chosen one. This change consists in filling the change region by the same spectral signature.
		\item Block Abundance: randomly select a region with the same spatial shape of the region affected by changes and replace original region abundances by the abundances of the second one. This produce a ``copy-paste'' pattern.
	\end{itemize}
Note that other change rules on the abundance matrix $\mathbf{A}^{t_1}$ could have been investigated; in particular some of them could require to include additional endmembers in the initial endmember matrix $\mathbf{M}^{t_1}$. The updated abundance $\mathbf{A}^{t_2}$ and endmember $\mathbf{M}^{t_2}$ matrices allow to define the after-change HR-HS latent image $\mathbf{X}^{t_2}$ as
\begin{equation*}
  \mathbf{X}^{t_2}=\mathbf{M}^{t_2}\mathbf{A}^{t_2}.
\end{equation*}
Fig. \ref{fig:CH} shows the four different change rules for one single selected region in image.

\subsubsection{Generating the observed images: spectral and spatial degradations}
\label{subsubsec:generating_observed}	
To produce spectrally degraded versions $\mathbf{Y}_{\mathrm{HR}}$ of the HR-HS latent image $\mathbf{X}^{t_j}$ ($j=1$ or $j=2$), two particular spectral responses have been used to assess the performance of the proposed algorithm when analyzing a HR-PAN or a $4$-band HR-MS image. The former has been obtained by uniformly averaging the first $43$ bands of the HR-HS pixel spectra. The later has been obtained by filtering the HR-HS latent image $\mathbf{X}^{t_j}$ by a $4$-band LANDSAT-like spectral response.

To generate a spatially degraded image, the HR-HS latent image $\mathbf{X}^{t_j}$ ($j=2$ or $j=1$) has been blurred by a $5 \times 5$ Gaussian kernel filter and down-sampled equally in vertical and horizontal directions with a factor $d = 5$. This spatial degradation operator implicitly relates the generated HR change mask $\mathbf{D}_{\mathrm{HR}}$ to its LR counterpart $\mathbf{D}_{\mathrm{LR}}$. Each LR pixel contains $d \times d$ HR pixels. As $\mathbf{D}_{\mathrm{HR}}$ is a binary mask, after the spatial degradation, if at least one of HR pixels associated to a given LR pixel is considered as a change pixel then the pixel in $\mathbf{D}_{\mathrm{LR}}$ is also considered as a change pixel.

\begin{figure}[h!]
\centering
%			\begin{subfigure}[b]{0.24\textwidth}
%					\centering	
%					\includegraphics[trim={3.08cm 2.35cm 3.08cm 1cm},clip,scale=0.255]{images/ref}
%					\caption{}
%					\label{fig:ref}
%			\end{subfigure}
			\begin{subfigure}[b]{0.24\textwidth}
					\centering	
					\includegraphics[trim={3.08cm 2.35cm 3.08cm 1cm},clip,scale=0.255]{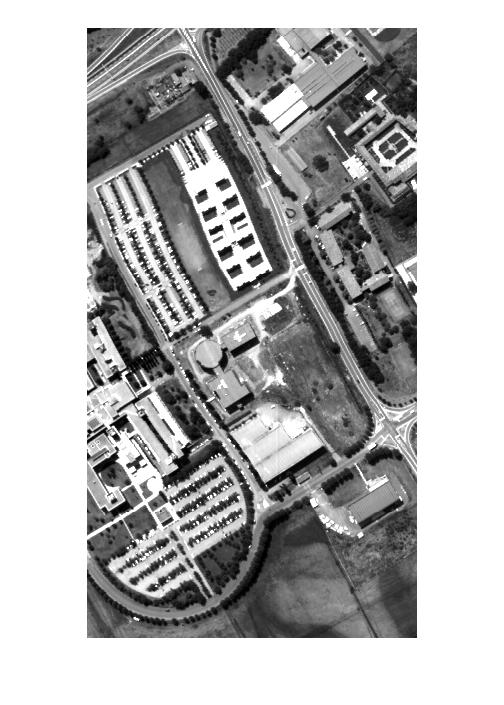}
					\caption{}
					\label{fig:panRef}
			\end{subfigure}
			\begin{subfigure}[b]{0.24\textwidth}
					\centering	
					\includegraphics[trim={3.08cm 2.35cm 3.08cm 1cm},clip,scale=0.255]{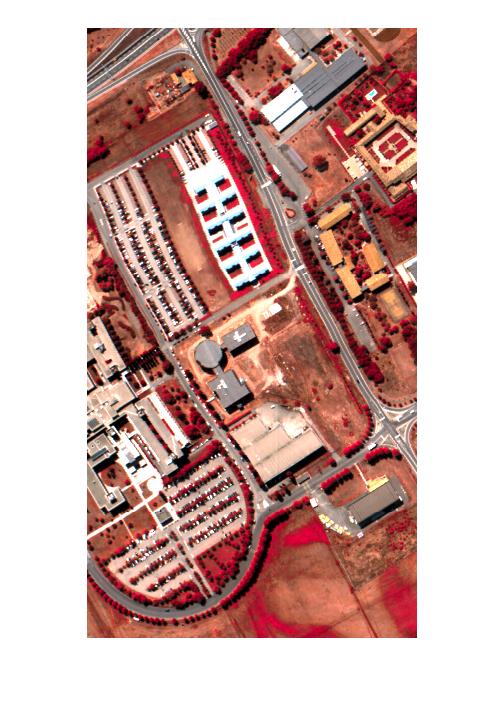}
					\caption{}
					\label{fig:msRef}
			\end{subfigure}
			\begin{subfigure}[b]{0.24\textwidth}
					\centering	
					\includegraphics[trim={3.08cm 2.35cm 3.08cm 1cm},clip,scale=1.27]{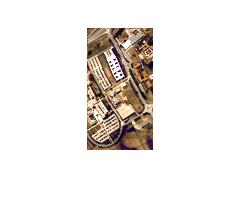}
					\caption{}
					\label{fig:hsRef}
				\end{subfigure}
			\caption{Degraded version of the reference image in Fig. \ref{fig:cref}: \protect\subref{fig:panRef} spectrally degraded HR-PAN image, \protect\subref{fig:msRef} spectrally degraded HR-MS image and \protect\subref{fig:hsRef} spatially degraded LR-HS image.}%
	\label{fig:DEG}%
\end{figure}

To illustrate the impact of these spectral and spatial degradations, Fig. \ref{fig:DEG} shows the HR-HS reference Pavia University image (a), corresponding HR-PAN (b) and HR-MS (c) images resulting from spectral degradations and a LR-HS image resulting from spatial degradation (d).

Note that, as mentioned in Section \ref{sec:CD}, the modality-time order can be arbitrary fixed, and without loss of generality, one may state either $t_1\leq t_2$ either $t_2\leq t_1$. Thus, there are $2$ distinct temporal configurations to generate the pair of observed HR and LR images:
\begin{itemize}
  \item Configuration 1: generating the spectrally (resp., spatially) degraded observed image $\mathbf{Y}_{\mathrm{HR}}$ (resp., $\mathbf{Y}_{\mathrm{LR}}$) from the before-change (resp., after-change) HR-HS latent image $\mathbf{X}^{t_1}$ (resp., $\mathbf{X}^{t_2}$),
  \item Configuration 2: generating the spectrally (resp., spatially) degraded observed image $\mathbf{Y}_{\mathrm{HR}}$ (resp., $\mathbf{Y}_{\mathrm{LR}}$) from the after-change (resp., before-change) HR-HS latent image $\mathbf{X}^{t_2}$ (resp., $\mathbf{X}^{t_1}$).
\end{itemize}

%Once the two HR-HS latent images $\mathbf{X}^{t_1}$ and $\mathbf{X}^{t_2}$ have been generated, two different scenarios can be considered. The first one uses $\mathbf{X}_{ref}$ to generate $\mathbf{Y}_{\mathrm{HR}}$ and $\mathbf{X}_{ch}$ to generate $\mathbf{Y}_{\mathrm{LR}}$ and the second one does the inverse. Note that for the proposed change detection framework there is not imposition of which input, $\mathbf{Y}_{\mathrm{HR}}$ or $\mathbf{Y}_{\mathrm{LR}}$, correspond to a changed scene since $\mathbf{X}_{ref}$ can also be seen as a changed version of $\mathbf{X}_{ch}$. Then, for a single pair $(\mathbf{X}_{ref},\mathbf{X}_{ch})$ it is possible to simulate $2$ different scenarios.

\subsection{Results}
\label{subsec:results}

The CD framework introduced in Section \ref{sec:CD} has been evaluated following the simulation protocol described in the previous paragraph. More precisely, $75$ regions have been randomly selected in the before-change HR-HS latent image $\mathbf{X}^{t_1}$ as those affected by changes. For each region, one of the three proposed change rules (zero-abundance, same abundance or block abundance) has been applied to build the after-change HR-HS latent image $\mathbf{X}^{t_2}$. The observed HR and LR images are generated according to one of the two temporal configurations discussed in Section \ref{subsubsec:generating_observed}. This leads to $150$ simulated pairs of HR-PAN/MS and LR-HS images. From each pair, as detailed in Section \ref{sec:CD}, one HR CD map $\hat{\mathbf{D}}_{\mathrm{HR}}$ and two LR CD maps $\hat{\mathbf{D}}_{\mathrm{LR}}$ and $\hat{\mathbf{D}}_{\mathrm{aLR}}$ are produced from the CD framework described in Fig. \ref{fig:CDF}. These HR and LR CD maps are respectively compared to the actual HR ${\mathbf{D}}_{\mathrm{HR}}$ and LR ${\mathbf{D}}_{\mathrm{LR}}$ masks to derive the empirical probabilities of false alarm $P_{\mathrm{FA}}$ and detection $P_{\mathrm{D}}$ that are represented as empirical receiver operating characteristics (ROC) curves, i.e., $P_{\mathrm{D}} = f(P_{\mathrm{FA}})$. These ROC curves have been averaged over the $150$ Monte Carlo simulations to mitigate the influence of time order and the influence of considered change region and rule.

Moreover, as quantitative figures-of-merit, two metrics derived from these ROC curves have been considered: i) the area under the curve (AUC), which is expected to be close to $1$ for a good testing rule and ii) a normalized distance between the no-detection point (defined by $P_{FA} = 1$ and  $P_{\mathrm{D}} = 0$) and the intersect of the ROC curve with the diagonal line $P_{\mathrm{FA}} = 1 - P_{\mathrm{D}}$, which should be close to $1$ for a good testing rule.

While implementing the proposed CD framework, the fusion step in Section \ref{subsec:fusion} has been conducted following the method proposed in \cite{weifast2015-2} with the Gaussian regularization because of its accuracy and computational efficiency. The corresponding regularization parameter has been chosen as $\lambda = 0.0001$ by cross-validation. Regarding the detection step, when considering multi-band images (i.e., MS or HS), the $4$ CD techniques detailed in Section \ref{subsec:fusion} (i.e., CVA, sCVA, MAD and IR-MAD) have been considered. Conversely, when considering PAN image, only CVA and sCVA have been considered since MAD and IR-MAD requires multi-band images. The sCVA method has been implemented with a window size of $L=7$ and $L={3,5,7}$ for PAN image.
%various window sizes $L$ according to the spatial resolution of the input images: $L \in \{3,7,13,21\}$ for LR-HS and $L \in \{3,13,21,43\}$ for HR-PAN/MS.

In absence of state-of-the-art CD techniques able to simultaneously handle images with distinct spatial and spectral resolutions, the proposed method has been compared to the crude approach that first consists in spatially (respectively spectrally) degrading the observed HR (respectively LR) image. The classical CD techniques described in Section \ref{sec:homCD} can then be applied to the resulting LR-MS/PAN images since they have the same, unfortunately low, spatial and spectral resolutions. The final result is a so-called worst-case LR CD mask denoted as $\hat{\mathbf{D}}_{\mathrm{WC}}$ in the following.

\begin{figure*}
    \centering
			\begin{subfigure}[b]{0.40\textwidth}
					\centering	
				  \includegraphics[trim={0.6cm 3.7cm 1.6cm 4.8cm},clip,width=0.8\textwidth]{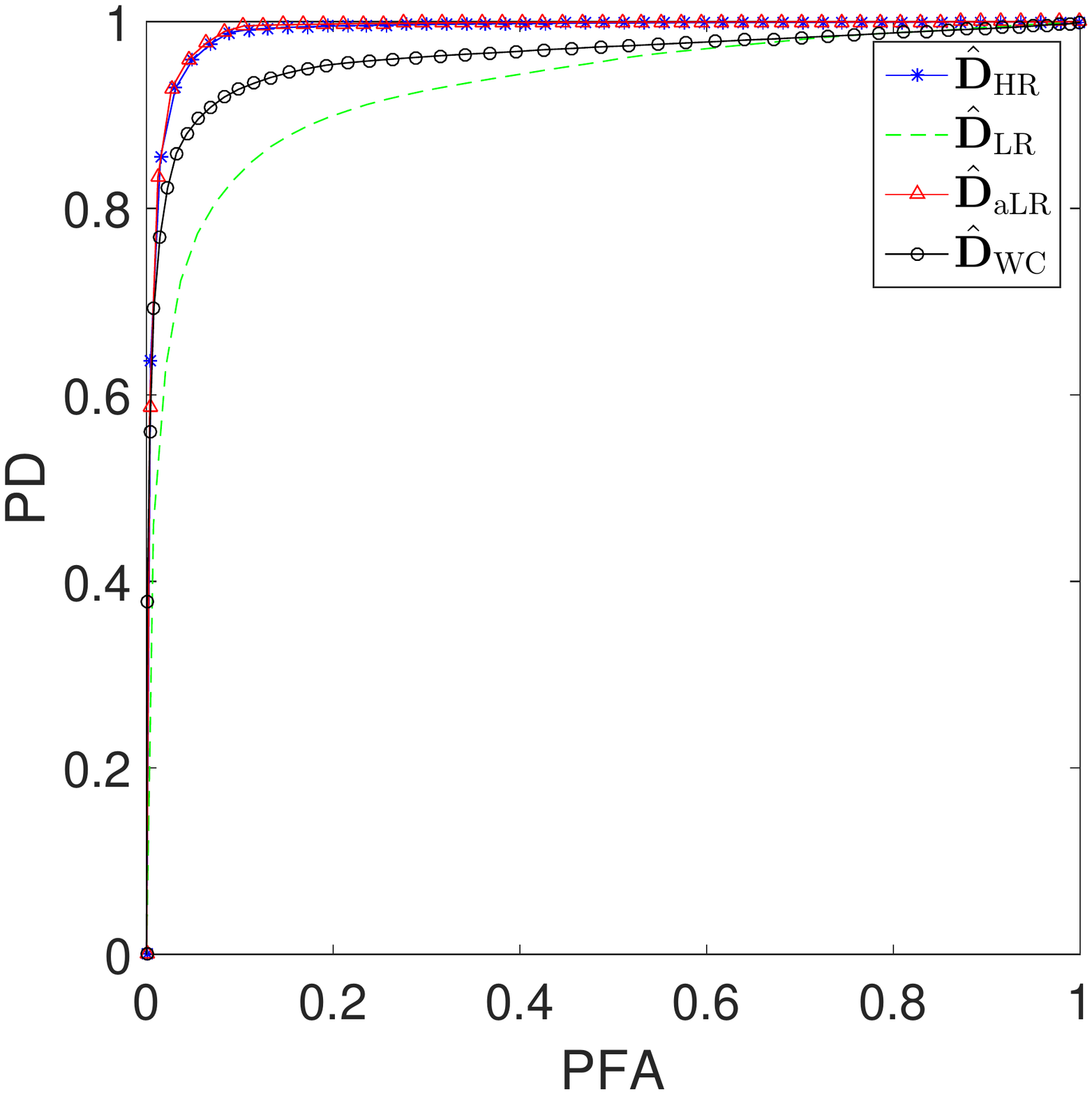}
					\caption{}
					\label{fig:rocCVA}
			\end{subfigure}
			\begin{subfigure}[b]{0.40\textwidth}
					\centering
				\includegraphics[trim={0.6cm 3.7cm 1.6cm 4.8cm},clip,width=0.8\textwidth]{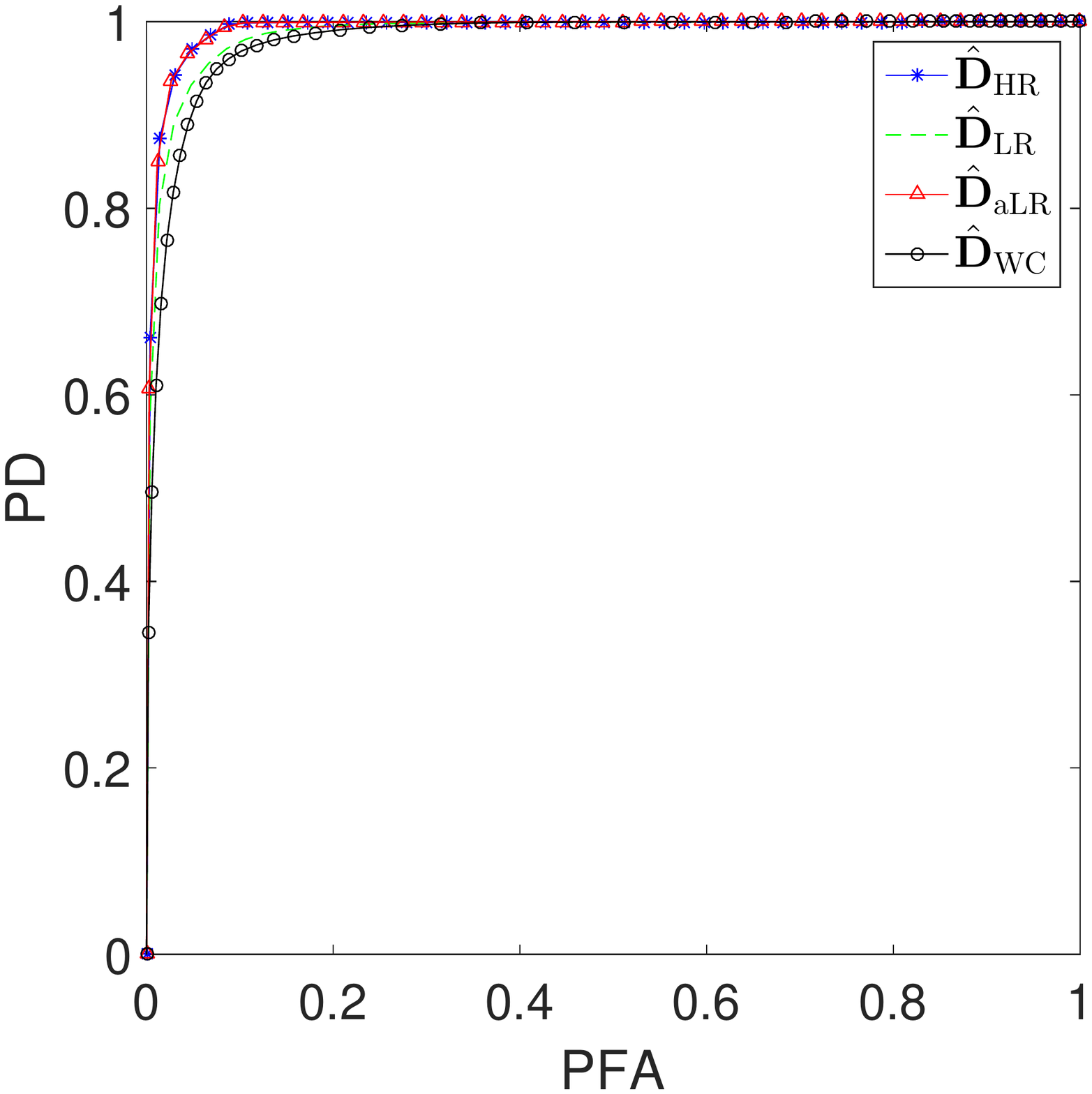}
					\caption{}
					\label{fig:rocsCVA7}
			\end{subfigure}
			\begin{subfigure}[b]{0.40\textwidth}
					\centering	
			\includegraphics[trim={0.6cm 3.7cm 1.6cm 4.8cm},clip,width=0.8\textwidth]{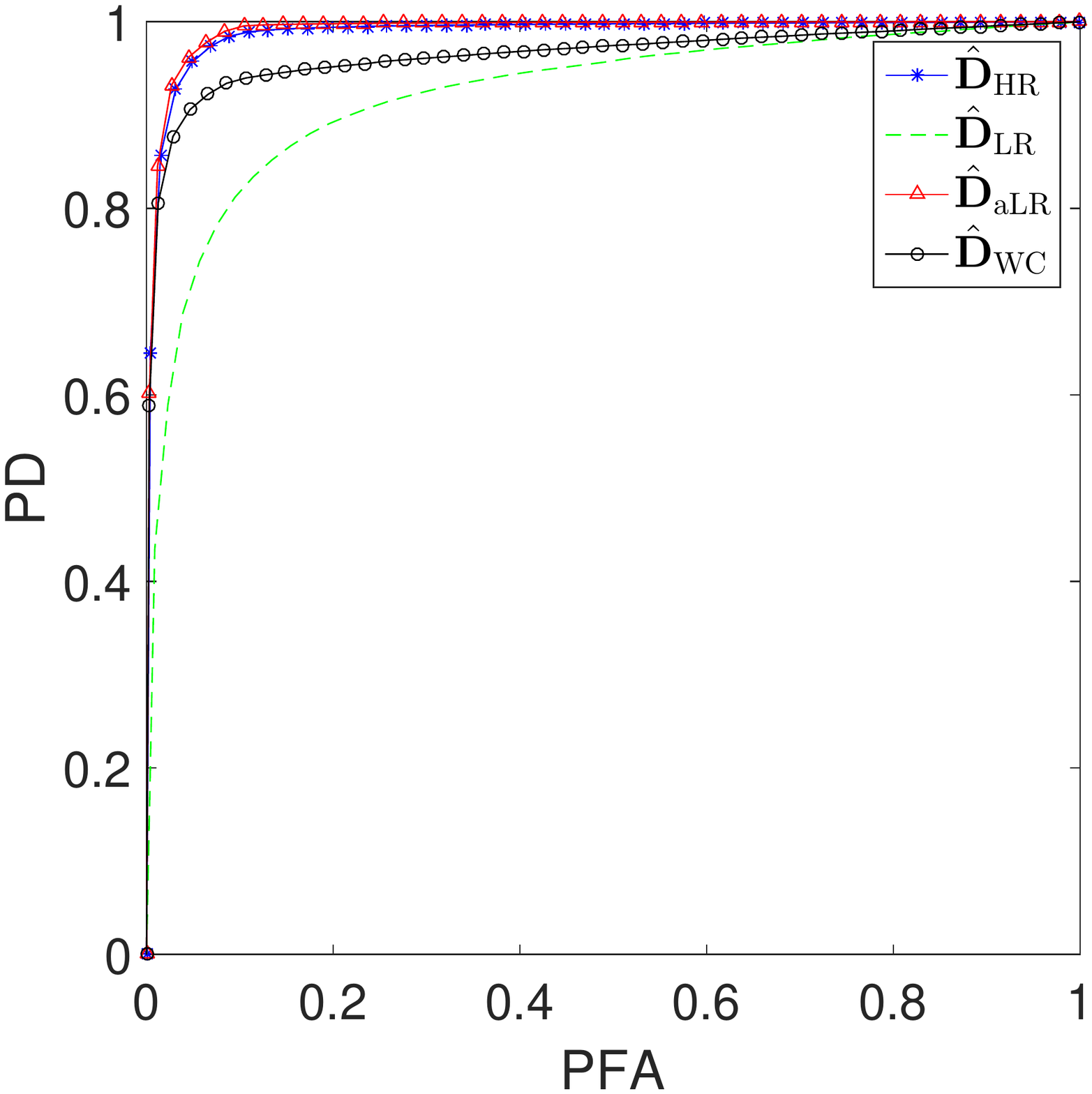}
					\caption{}
					\label{fig:rocMAD}
			\end{subfigure}
			\begin{subfigure}[b]{0.40\textwidth}
					\centering	
					\includegraphics[trim={0.6cm 3.7cm 1.6cm 4.8cm},clip,width=0.8\textwidth]{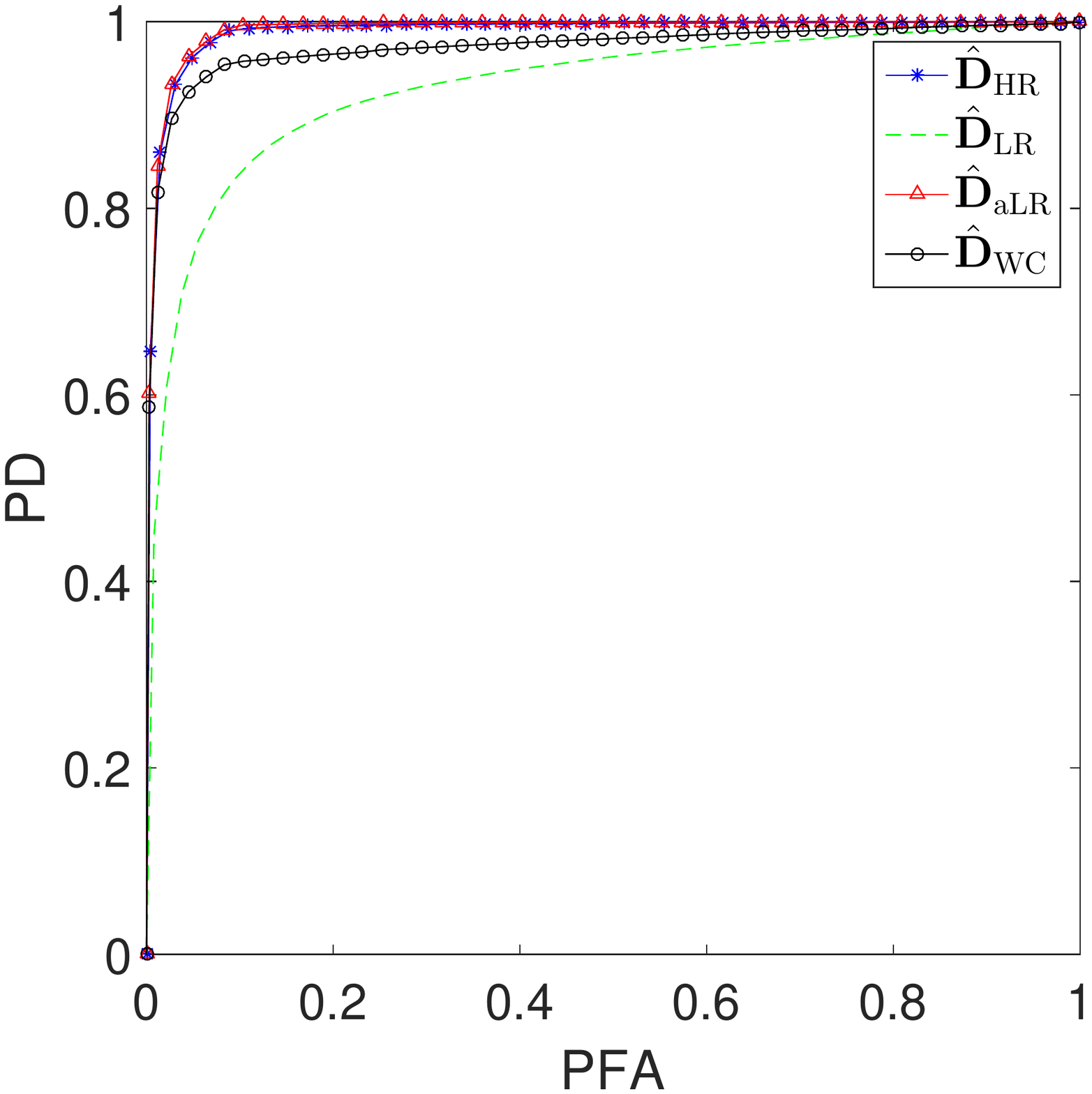}
					\caption{}
					\label{fig:rocIRMAD}
			\end{subfigure}
			\caption{Scenario $1$: ROC curves computed from \protect\subref{fig:rocCVA} CVA, \protect\subref{fig:rocsCVA7} sCVA(7), \protect\subref{fig:rocMAD} MAD and \protect\subref{fig:rocIRMAD} IRMAD. \label{fig:ROCSEN1}}%
\end{figure*}

\subsubsection{Scenario $1$: Change detection between HR-MS and LR-HS images}

The first simulation scenario considers a set of HR-MS and LR-HS images. The ROC curves are plotted in Fig. \ref{fig:ROCSEN1} with corresponding performance metrics reported in Table \ref{table:ROCSEN1}. These results show that, whatever the implemented CD testing feature (CVA, sCVA, MAD or IR-MAD), the proposed framework offers high precision. In particular, the aLR change map $\hat{\mathbf{A}}_{\mathrm{aLR}}$ computed from the estimated HR change map $\hat{\mathbf{D}}_{\mathrm{HR}}$ provides significantly better results that those obtained in the worst-case and those obtained on the estimated LR change map $\hat{\mathbf{D}}_{\mathrm{LR}}$ directly. This can be explained by the intrinsic quality of the estimated HR change map $\hat{\mathbf{D}}_{\mathrm{HR}}$, which roughly provides similar detection performance as the aLR change map $\hat{\mathbf{D}}_{\mathrm{aLR}}$ with the great advantage to be available at a finer spatial resolution.

\newcommand{\one}[1]{\bf{\textcolor[rgb]{0.00,0.00,1.00}{#1}}}
\newcommand{\two}[1]{\textcolor[rgb]{0.00,0.00,1.00}{#1}}
\setlength{\tabcolsep}{5pt}
\renewcommand{\arraystretch}{1.3}

\begin{table}[h!]
\caption{Scenario $1$: detection performance in terms of AUC and normalized distance.}
\label{table:ROCSEN1}
\centering
\begin{tabular}{|c|c|c|c|c|c|}
\cline{3-6}
\multicolumn{2}{c|}{}      & {$\hat{\mathbf{D}}_{\mathrm{HR}}$} & $\hat{\mathbf{D}}_{\mathrm{LR}}$ & $\hat{\mathbf{D}}_{\mathrm{aLR}}$ & $\hat{\mathbf{D}}_{\mathrm{WC}}$\\
\hline
\hline
\multirow{2}{*}{\rotatebox{00}{CVA}}          & AUC   & $\two{0.988800}$ & $0.928687$& $\one{0.990373}$ & $0.961285$\\
                                              & Dist. & $\two{0.953895}$ & $0.866487$& $\one{0.956896}$ & $0.918192$\\
\hline
\multirow{2}{*}{\rotatebox{00}{sCVA($7$)}}    & AUC   & $\two{0.991916}$ & $0.986532$& $\one{0.992090 }$ & $0.980919$\\
                                              & Dist. & $\two{0.958996}$ & $0.943194$& $\one{0.959396}$ & $0.935594$\\
\hline
\multirow{2}{*}{\rotatebox{00}{MAD}}          & AUC   & $\two{0.988032}$ & $0.922616$& $\one{0.990971}$ & $0.964098$\\
                                              & Dist. & $\two{0.953095}$ & $0.857786$& $\one{0.957696}$ & $0.926693$\\
\hline
\multirow{2}{*}{\rotatebox{00}{IR-MAD}}       & AUC   & $\two{0.989237}$ & $0.929343$& $\one{0.991151}$ & $0.972520$\\
                                              & Dist. & $\two{0.954995}$ & $0.867587$& $\one{0.958096}$ & $0.938594$\\
\hline
\end{tabular}
\end{table}

%Figure \ref{fig:MSHSMSLR} and Table \ref{table:MSHSMSLR} present the result of the lowest resolution degradation ROC for no-noise observation images. Comparing such results with those of Figure \ref{fig:MSHSMSMSLR} shows that, in the worst spatial resolution case, the proposed method offers the highest precision. Now, comparing with Figure \ref{fig:MSHSMS} it is obvious that the proposed method offers a higher precision in a finer resolution.

\begin{figure*}
\centering
			\begin{subfigure}[b]{0.24\textwidth}
					\centering	
					\includegraphics[trim={2.3cm 1.42cm 2.3cm 0.7cm},clip,scale=0.32]{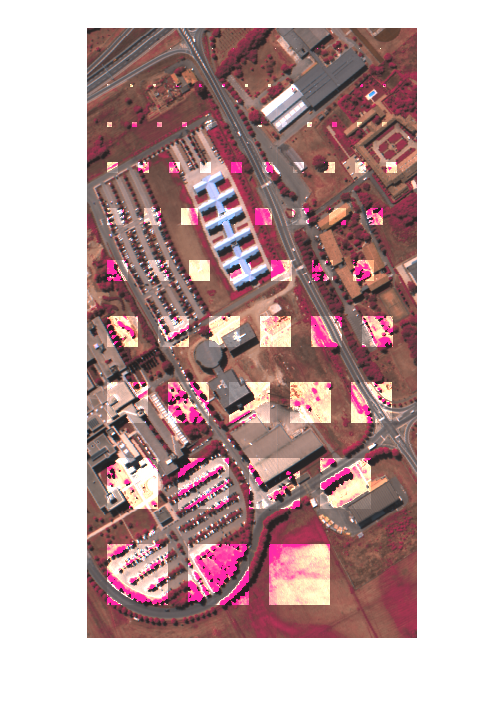}
					\caption{$\mathbf{Y}_{\mathrm{HR}}$}
					\label{fig:obsMS}
			\end{subfigure}
			\begin{subfigure}[b]{0.24\textwidth}
					\centering	
					\includegraphics[trim={2.3cm 1.7cm 2.3cm 0.7cm},clip,scale=1.6]{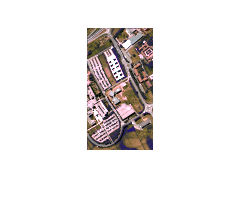}
					\caption{$\mathbf{Y}_{\mathrm{LR}}$}
					\label{fig:obsHS}
			\end{subfigure}
			\begin{subfigure}[b]{0.24\textwidth}
					\centering	
					\includegraphics[trim={2.3cm 1.42cm 2.3cm 0.7cm},clip,scale=0.32]{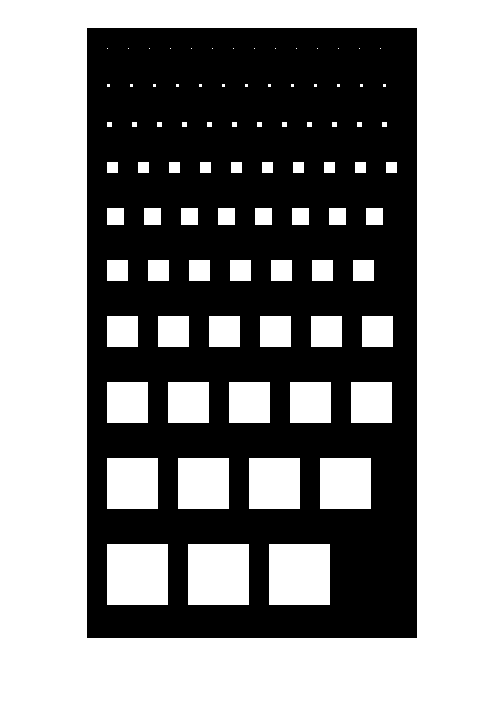}
					\caption{$\mathbf{D}_{\mathrm{HR}}$}
					\label{fig:maskHighRes}
			\end{subfigure}
			\begin{subfigure}[b]{0.24\textwidth}
					\centering	
					\includegraphics[trim={2.3cm 1.7cm 2.3cm 0.7cm},clip,scale=1.6]{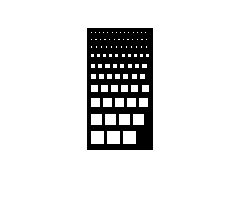}
					\caption{$\mathbf{D}_{\mathrm{LR}}$}
					\label{fig:maskLowRes}
				\end{subfigure}
						\begin{subfigure}[b]{0.24\textwidth}
					\centering	
					\includegraphics[trim={2.3cm 1.42cm 2.3cm 0.7cm},clip,scale=0.32]{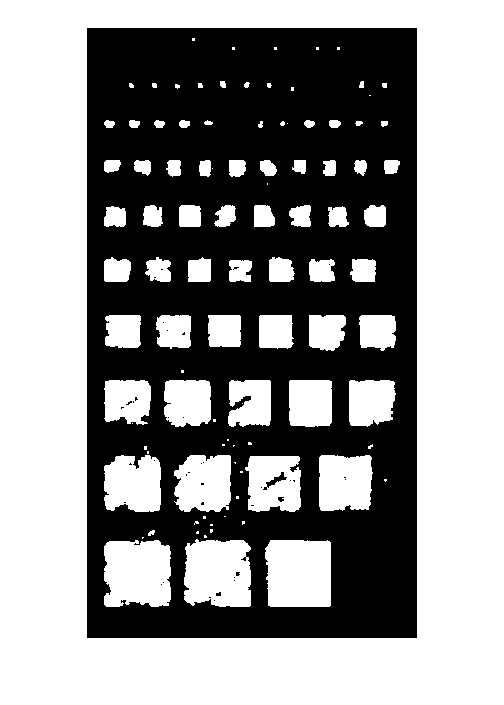}
					\caption{$\hat{\mathbf{D}}_{\mathrm{HR}}$}
					\label{fig:changeDetectionHR}
			\end{subfigure}
			\begin{subfigure}[b]{0.24\textwidth}
					\centering	
					\includegraphics[trim={15cm 10cm 2.3cm 0.7cm},clip,scale=0.275]{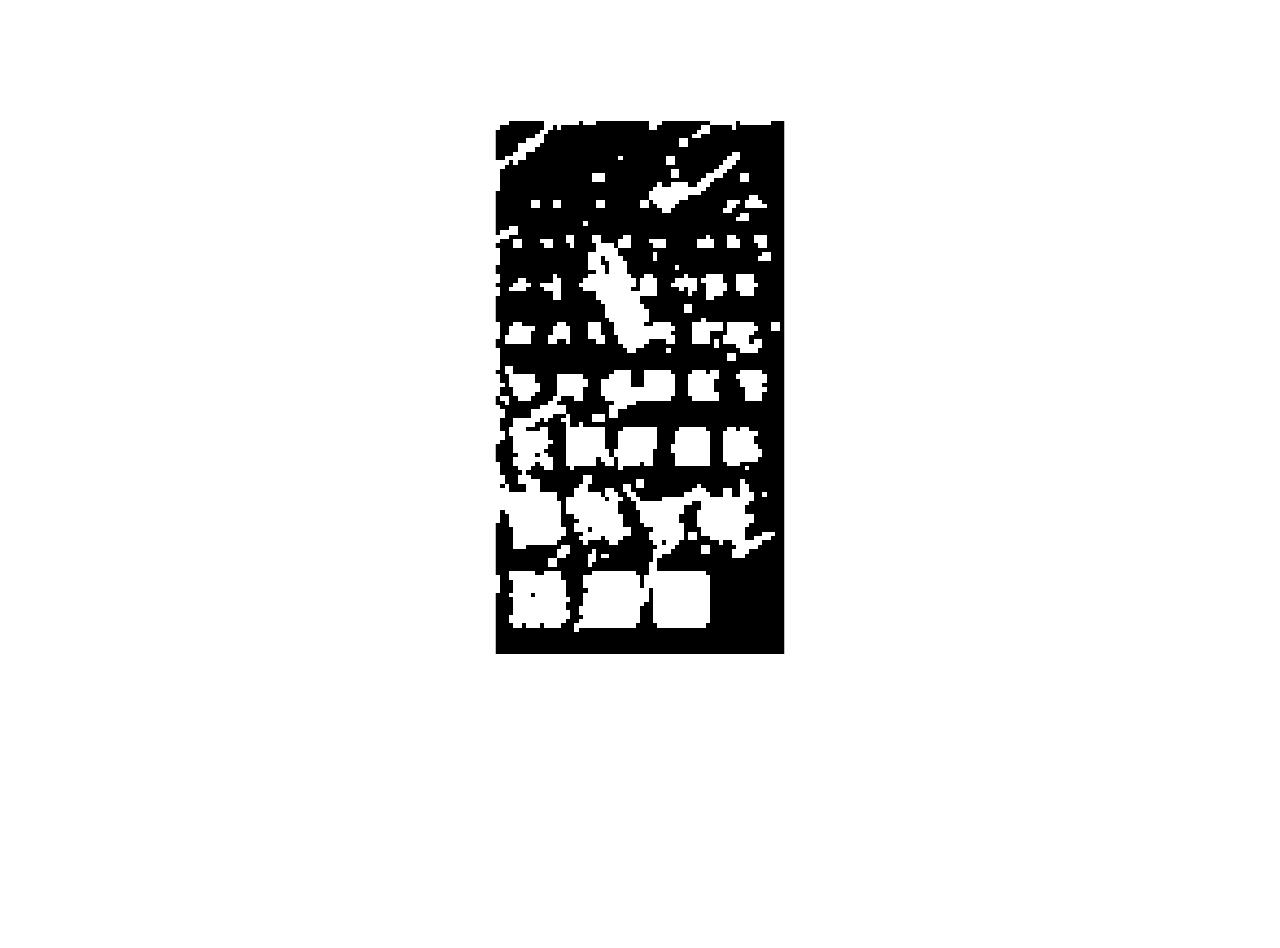}
					\caption{$\hat{\mathbf{D}}_{\mathrm{LR}}$}
					\label{fig:changeDetectionLR}
			\end{subfigure}
			\begin{subfigure}[b]{0.24\textwidth}
					\centering	
					\includegraphics[trim={2.3cm 1.7cm 2.3cm 0.7cm},clip,scale=1.6]{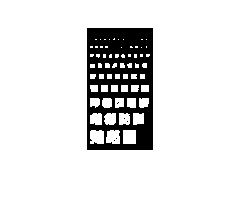}
					\caption{$\hat{\mathbf{D}}_{\mathrm{aLR}}$}
					\label{fig:changeDetectionMSLR}
			\end{subfigure}
			\begin{subfigure}[b]{0.24\textwidth}
					\centering	
					\includegraphics[trim={2.3cm 1.7cm 2.3cm 0.7cm},clip,scale=1.6]{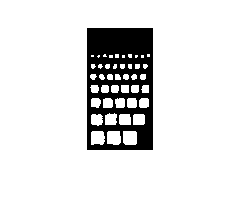}
					\caption{$\hat{\mathbf{D}}_{\mathrm{WC}}$}
					\label{fig:changeDetectionWC}
			\end{subfigure}
			\caption{Scenario $1$: \protect\subref{fig:obsMS} observed HR-MS image, \protect\subref{fig:obsHS} observed LR-HS image, \protect\subref{fig:maskHighRes} actual HR CD mask $\mathbf{D}_{\mathrm{HR}}$, \protect\subref{fig:maskLowRes} actual LR CD mask $\mathbf{D}_{\mathrm{LR}}$, \protect\subref{fig:changeDetectionHR} estimated HR CD map $\hat{\mathbf{D}}_{\mathrm{HR}}$, \protect\subref{fig:changeDetectionHR} estimated LR CD map $\hat{\mathbf{D}}_{\mathrm{LR}}$, \protect\subref{fig:changeDetectionMSLR} estimated aLR CD map $\hat{\mathbf{D}}_{\mathrm{aLR}}$ and \protect\subref{fig:changeDetectionWC} worst-case CD map $\hat{\mathbf{D}}_{\mathrm{WC}}$. \label{fig:precision}}%
\end{figure*}

To visually illustrate this finding, Fig. \ref{fig:precision} shows the CD maps estimated from a pair of observed HR-MS \subref{fig:obsMS} and LR-HS \subref{fig:obsHS} images containing multiple changes with size varying from $1\times 1$-pixel to $61 \times 61$-pixels using sCVA(3) classical CD. The actual HR and LR CD masks are reported in Fig. \ref{fig:precision}\subref{fig:maskHighRes} and \subref{fig:maskLowRes}, respectively. Figures \ref{fig:precision}\subref{fig:changeDetectionHR} to \subref{fig:changeDetectionWC} show the estimated CD maps $\hat{\mathbf{D}}_{\mathrm{HR}}$, $\hat{\mathbf{D}}_{\mathrm{LR}}$, $\hat{\mathbf{D}}_{\mathrm{aLR}}$ and $\hat{\mathbf{D}}_{\mathrm{WC}}$, respectively. % for a threshold chosen to get the best cut-off selection of ROC curve where $PFA = 1 - PD = PND$.
Once again, these results clearly demonstrate that the HR CD map $\hat{\mathbf{D}}_{\mathrm{HR}}$ estimated by the proposed method achieves a better detection rate with a higher precision.

\subsubsection{Scenario $2$: Change detection between HR-PAN and LR-HS images}

\begin{figure*}
        \centering
			\begin{subfigure}[b]{0.40\textwidth}
					\centering	
				\includegraphics[trim={0.6cm 3.7cm 1.6cm 4.8cm},clip,width=0.8\textwidth]{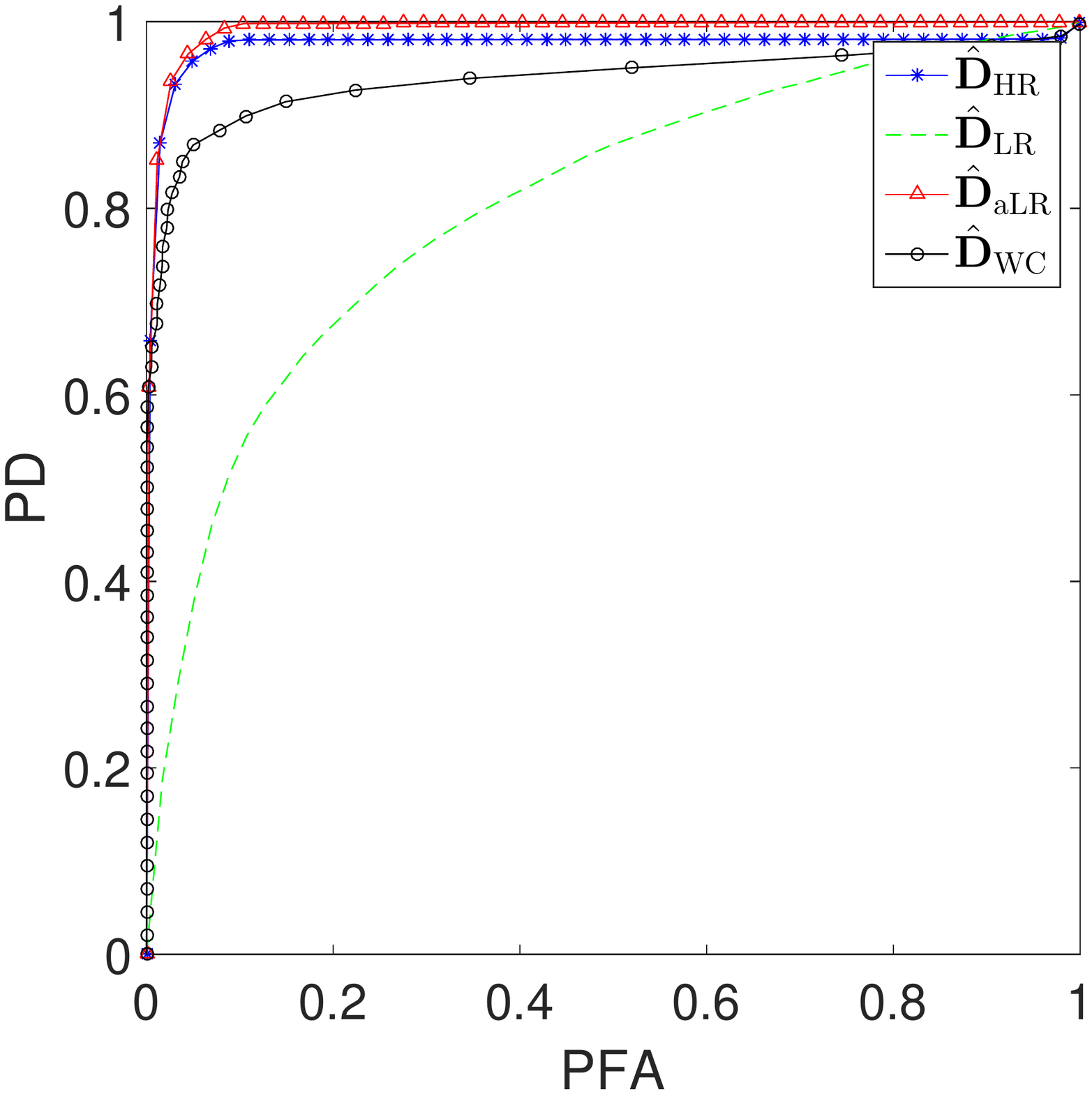}
					\caption{}
					\label{fig:CVAPAN}
			\end{subfigure}
			\begin{subfigure}[b]{0.40\textwidth}
					\centering
			    \includegraphics[trim={0.6cm 3.7cm 1.6cm 4.8cm},clip,width=0.8\textwidth]{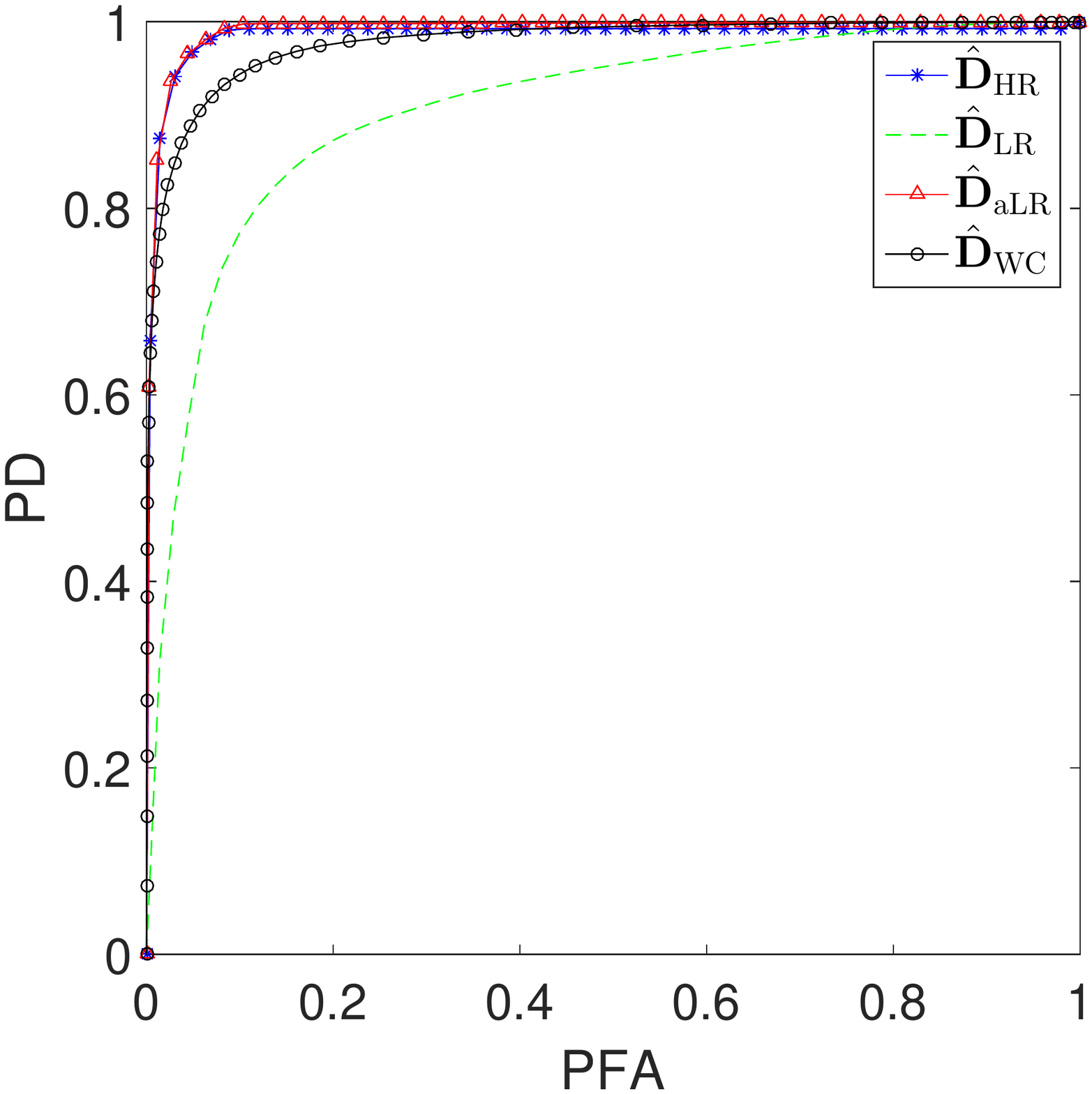}
					\caption{}
					\label{fig:sCVA3PAN}
			\end{subfigure}
			\begin{subfigure}[b]{0.40\textwidth}
					\centering	
				\includegraphics[trim={0.6cm 3.7cm 1.6cm 4.8cm},clip,width=0.8\textwidth]{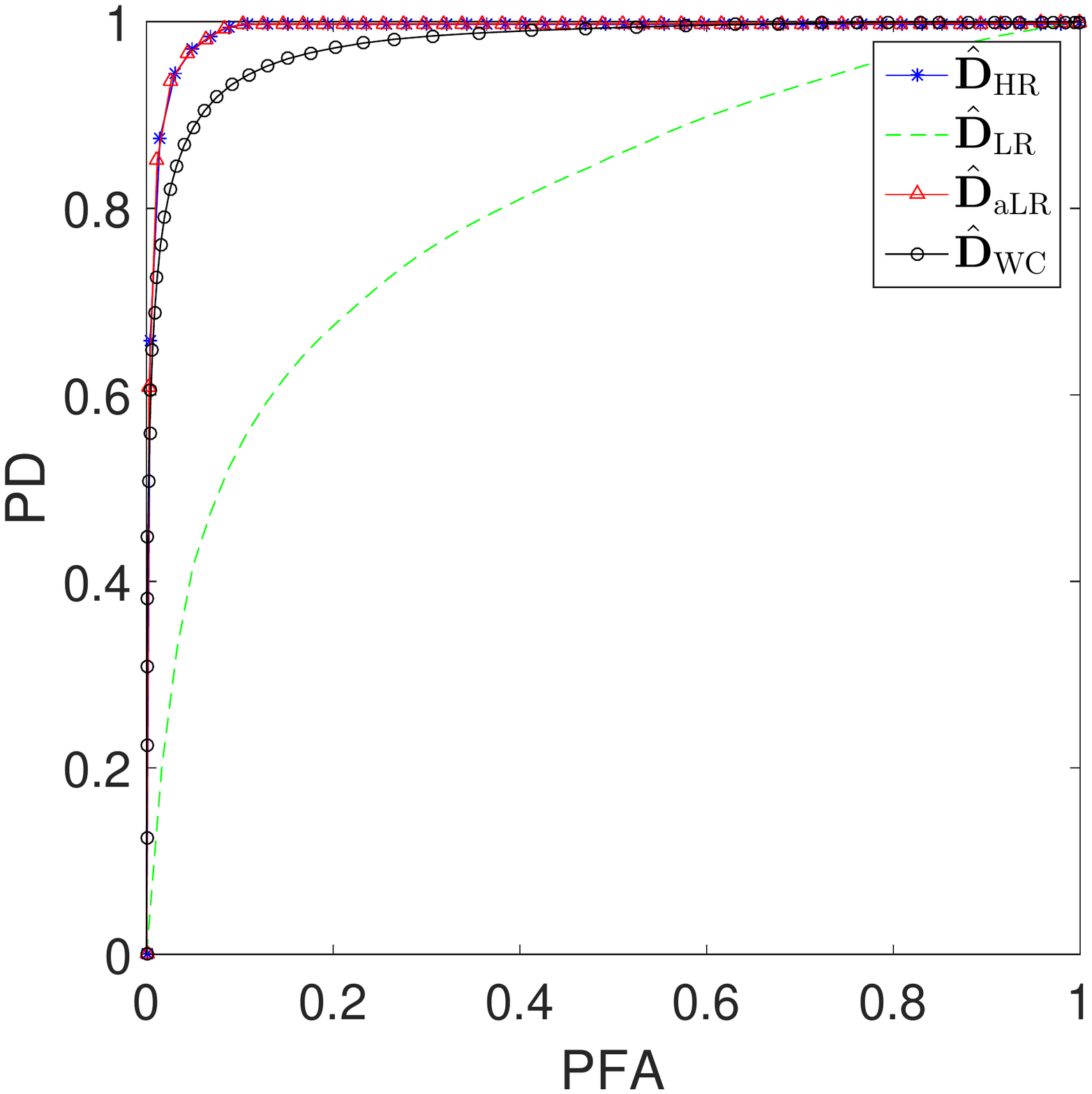}
					\caption{}
					\label{fig:sCVA5PAN}
			\end{subfigure}
			\begin{subfigure}[b]{0.40\textwidth}
					\centering
			\includegraphics[trim={0.6cm 3.7cm 1.6cm 4.8cm},clip,width=0.8\textwidth]{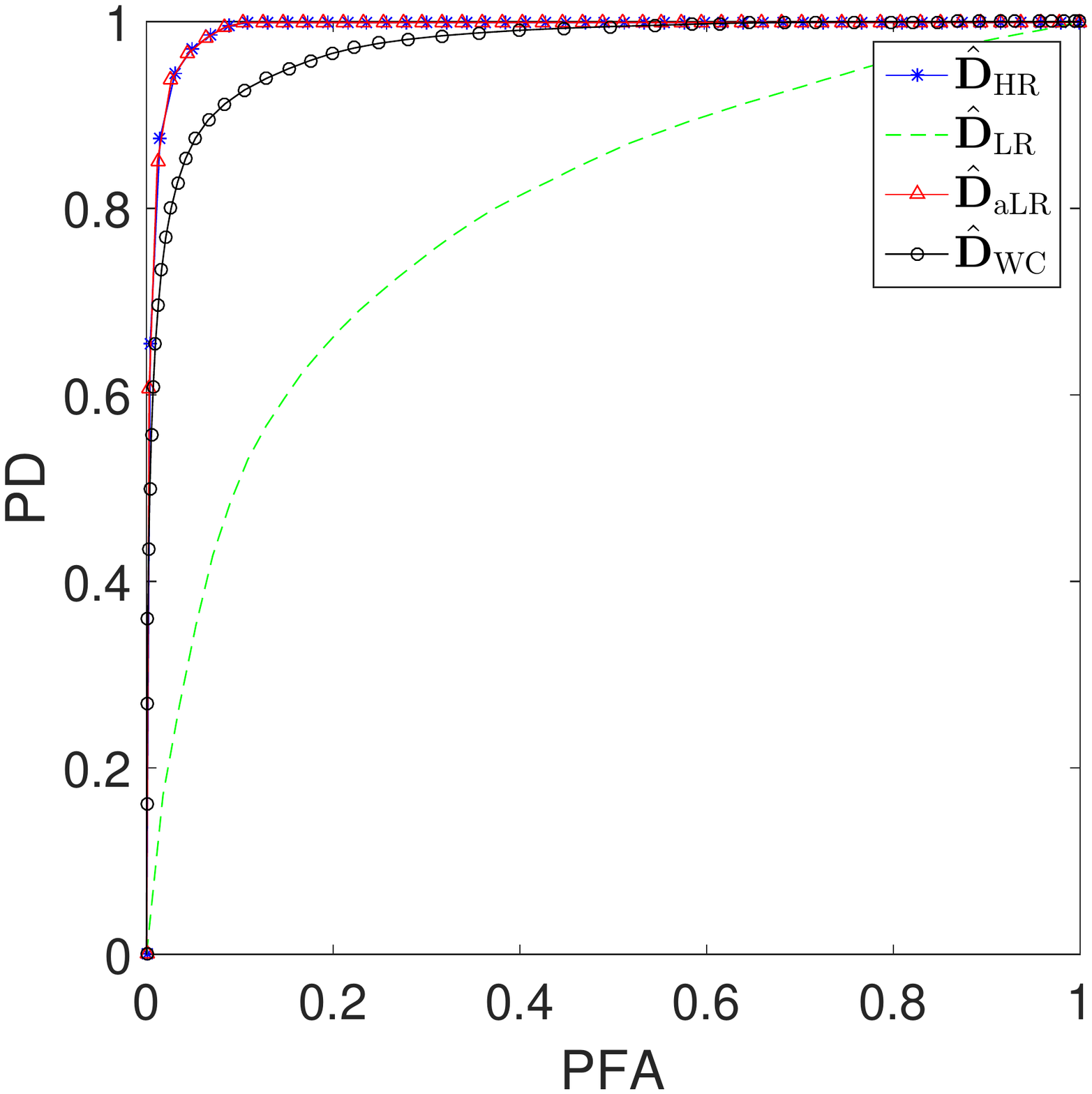}
					\caption{}
					\label{fig:sCVA7PAN}
			\end{subfigure}
			\caption{Scenario 2: ROC curves computed from  \protect\subref{fig:CVAPAN} CVA, \protect\subref{fig:sCVA3PAN} sCVA(3), \protect\subref{fig:sCVA5PAN} sCVA(5) and \protect\subref{fig:sCVA7PAN} sCVA(7). \label{fig:ROCSEN2}}%
\end{figure*}

In the second scenario, the same procedure as Scenario $1$ has been considered while replacing the observed MS image by a PAN image. The ROC curves are depicted in Fig. \ref{fig:ROCSEN2} with corresponding metrics in Table \ref{table:ROCSEN2}. As for Scenario $1$, whatever the decision technique (CVA or its spatially regularized counterpart sCVA), the comparison of these curves show that the HR CD map also leads to a higher accuracy, since it is sharper than the LR maps. In particular, it provides a significantly more powerful test than the crude approach that consists in degrading both observed HR-PAN and LR-HS images to reach the same spatial and spectral resolutions.

%As in Scenario 1, comparing Figure \ref{fig:PANHSPANLR} with Figure \ref{fig:PANHSPAN} and Figure \ref{fig:PANHSPANMSLR} it is possible to conclude that the proposed method offers a higher precision in a both resolutions even with a lower spectral resolution of HR observed image.

%Figure \ref{fig:PANHSPAN} with Table \ref{table:PANHSPAN} and Figure \ref{fig:PANHSPAN20dB} with table \ref{table:PANHSPAN20dB} present the ROC results for a couple HR-PAN and LR-HS of observed images investigating in the PAN spatial resolution for no-noise and $SNR = 20dB$ respectively. Figure \ref{fig:PANHSPANMSLR} and Table \ref{table:PANHSPANMSLR} shows the ROC results corresponding to the LR map, using the HR map downsampled by $\mathbf{S}$.

\begin{table}[h!]
    \caption{Scenario 2: detection performance in terms of AUC and normalized distance.}
    \centering
    \begin{tabular}{|c|c|c|c|c|c|}
    \cline{3-6}
    \multicolumn{2}{c|}{}      & $\hat{\mathbf{D}}_{\mathrm{HR}}$ & $\hat{\mathbf{D}}_{\mathrm{LR}}$ & $\hat{\mathbf{D}}_{\mathrm{aLR}}$ & $\hat{\mathbf{D}}_{\mathrm{WC}}$\\
    \hline
    \hline
    \multirow{2}{*}{\rotatebox{00}{CVA}}            & AUC   & $\two{0.973951}$ & $0.802892$& $\one{0.991192}$ & $0.937878$\\
                                                    & Dist. & $\two{0.953595}$ & $0.734773$& $\one{0.959296}$ & $0.897090$\\
    \hline
    \multirow{2}{*}{\rotatebox{00}{sCVA($3$)}}      & AUC   & $\two{0.985251}$ &0.906041& $\one{0.991008}$ & $0.977657$\\
                                                    & Dist. & $\two{0.957796}$ &0.842784& $\one{0.959296}$ & $0.924892$\\
    \hline
    \multirow{2}{*}{\rotatebox{00}{sCVA($5$)}}      & AUC   & $\two{0.989493}$ & $0.801926$& $\one{0.990435}$ & $0.975165$\\
                                                    & Dist. & $\two{0.958996}$ & $0.732073$& $\one{0.959196}$ & $0.921792$\\
    \hline
    \multirow{2}{*}{\rotatebox{00}{sCVA($7$)}}      & AUC   & $\two{0.991175}$ & $0.794184$& $\one{0.991545}$ & $0.971988$\\
                                                    & Dist. & $\two{0.959296}$ & $0.728073$& $\one{0.959496}$ & $0.913691$\\
    \hline
    \end{tabular}
        \label{table:ROCSEN2}
\end{table}

\section{Conclusions and future Work}
\label{sec:conclusion}
This paper introduced an unsupervised change detection framework for handling multi-band optical images of different modalities, i.e., with different spatial and spectral resolutions. The method was based on a $3$-step procedure. The first step performed the fusion of the two different spatial/spectral resolution multi-band optical images to recover a pseudo-latent image of high spatial and spectral resolutions. From this fused image, the second step generated a pair of predicted images with the same resolutions as the observed multi-band images. Finally, standard CD techniques were applied to each pair of observed and predicted images with same spatial and spectral resolutions. The relevance of the proposed framework was assessed thanks to an experimental protocol. These experiments demonstrated the accuracy of the recovered high-resolution change detection map.

Future works will include the generalization of the proposed framework to deal with images of other modalities. Indeed, the newly proposed $3$-step procedure (\emph{fusion}, \emph{prediction}, \emph{detection}) is expected to be applicable provided that a physically-based direct model can be derived to relate the observed images with a pseudo-latent image.

\bibliographystyle{ieeetran}
\bibliography{strings_all_ref,HSbib_cleaned}

\end{document}